\documentclass[a4paper, 10pt, conference]{ieeeconf} 
\IEEEoverridecommandlockouts  

\makeatletter
\let\NAT@parse\undefined
\makeatother
\usepackage{natbib}
\usepackage{ upgreek }
\usepackage{verbatim} 
\usepackage{tikz} 

\usepackage{amssymb}
\usepackage{graphicx}
\usepackage{url}
\usepackage{verbatim} 
\usepackage{color}
\usepackage{array}
\usepackage{relsize}
\usepackage{multirow}
\usepackage{amsmath}

\usepackage{hyperref}
\DeclareMathOperator*{\argmax}{argmax}

\long\def\symbolfootnote[#1]#2{\begingroup%
\def\thefootnote{\fnsymbol{footnote}}\footnotetext[#1]{#2}\endgroup}
\newlength\savedwidth
\newcommand\whline[1]{\noalign{\global\savedwidth\arrayrulewidth
                               \global\arrayrulewidth #1} %
                      \hline
                      \noalign{\global\arrayrulewidth\savedwidth}}

\newcommand{\todo}[1]{\textcolor{blue}{\textbf{#1}}}

\newcommand{\n}{{n}}             
\newcommand{\x}{{\mathbf x}}     
\newcommand{\xs}[1]{{x_{#1}}}    
\newcommand{\y}{{\mathbf y}}     
\newcommand{\ys}[1]{{y_{#1}}}    
\newcommand{\ysc}[2]{{y_{#1}^{#2}}}    
\newcommand{\zsc}[2]{{z_{#1}^{#2}}}    
\newcommand{\fn}[1]{{\phi_n(#1)}}      
\newcommand{\fe}[3]{{\phi_{#1}(#2,#3)}}
\newcommand{\fnp}[1]{{\phi_n'(#1)}}      
\newcommand{\fep}[3]{{\phi_{#1}'(#2,#3)}}
\newcommand{\w}{{\mathbf w}}           
\newcommand{\wn}[1]{{w_n^{#1}}}        
\newcommand{\wnp}[1]{{w_n'^{#1}}}        
\newcommand{\we}[3]{{w_{#1}^{#2#3}}}   
\newcommand{\wep}[3]{{w_{#1}'^{#2\;#3}}}   
\newcommand{\df}[3]{{f_{#3}(#1,#2)}}   
\newcommand{\loss}[2]{{\Delta(#1,#2)}}   

\title{\LARGE \bf
Contextually Guided Semantic Labeling and Search for 3D Point Clouds
}
\author{Abhishek Anand$^*$, Hema Swetha Koppula$^*$, Thorsten Joachims, Ashutosh Saxena\\
Department of Computer Science, Cornell University.\\
\tt{\{aa755,hema,tj,asaxena\}@cs.cornell.edu}}

\begin{document}
\maketitle

\begin{abstract}

RGB-D cameras, which give an RGB image together with depths,
are becoming increasingly popular for robotic perception.
In this paper, we address the task of detecting commonly found objects 
in the 3D point cloud of indoor scenes obtained from such cameras. 
Our method uses a graphical model that captures various features and contextual
relations, including the local visual appearance and 
 shape cues, object co-occurence relationships and geometric relationships.
With a large number of object classes and relations, the model's parsimony
becomes important and we address that by using multiple types of edge
potentials.  We train the model using a maximum-margin learning approach.  In
our experiments over a total of 52 3D scenes of homes and offices (composed
from about 550 views), we get a performance of 84.06\% and 73.38\% in labeling 
office and home scenes respectively for 17 object classes each.
We also present a method for a robot to search for an object 
using the learned model and the contextual information available from the current labelings of the scene.
We applied this algorithm successfully on a mobile robot for the 
task of finding 12 object classes in 10 different offices and achieved a precision of 97.56\% with 78.43\% recall.\footnote{Parts of this work have been published at \citep{nips:3dlabeling,rgbd:3dlabeling};  those works did not present our contextually-guided search algorithm and the robotic experiments. This submission also includes more details on the algorithm and results.}

\end{abstract}

\symbolfootnote[0]{$^*$ indicates equal contribution.}

\section{Introduction}

Inexpensive RGB-D sensors that augment an RGB image with depth data have recently 
become widely available. These cameras are increasingly becoming the de-facto standard for perception for many robots. 
At the same time, years of research on SLAM (Simultaneous Localization and Mapping) has now
made it possible to merge multiple RGB-D images into a single point cloud, easily providing an approximate 3D model of a complete indoor scene (i.e., a room). In this paper, we explore how this move from part-of-scene 2D images to full-scene 3D point clouds can 
improve the richness of models for object labeling.



In the past, a significant amount of work has been done in semantic labeling of 2D 
images \citep{Torralba:exploting_context, HeitzECCV_usingstufftofindthings, felzenszwalb2008discriminatively, Collet2011,li2011feccm}. However, a  lot of valuable information about the 3D shape and geometric layout of objects is lost when a 2D image is formed from the corresponding 3D world. A classifier that has access to a full 3D model can access 
important geometric properties in addition to the local shape and appearance of an object.  For example, 
many objects occur in characteristic relative geometric configurations (e.g., a monitor is almost always on a table), and many objects consist of visually distinct parts that occur in a certain relative configuration. More generally, a 3D model makes it possible to reason about a variety of 3D properties such as 
3D distances, volume and local convexity.
 
    \begin{figure}[t!]
 \centering
\includegraphics[width=.20\linewidth,height=1.8in]{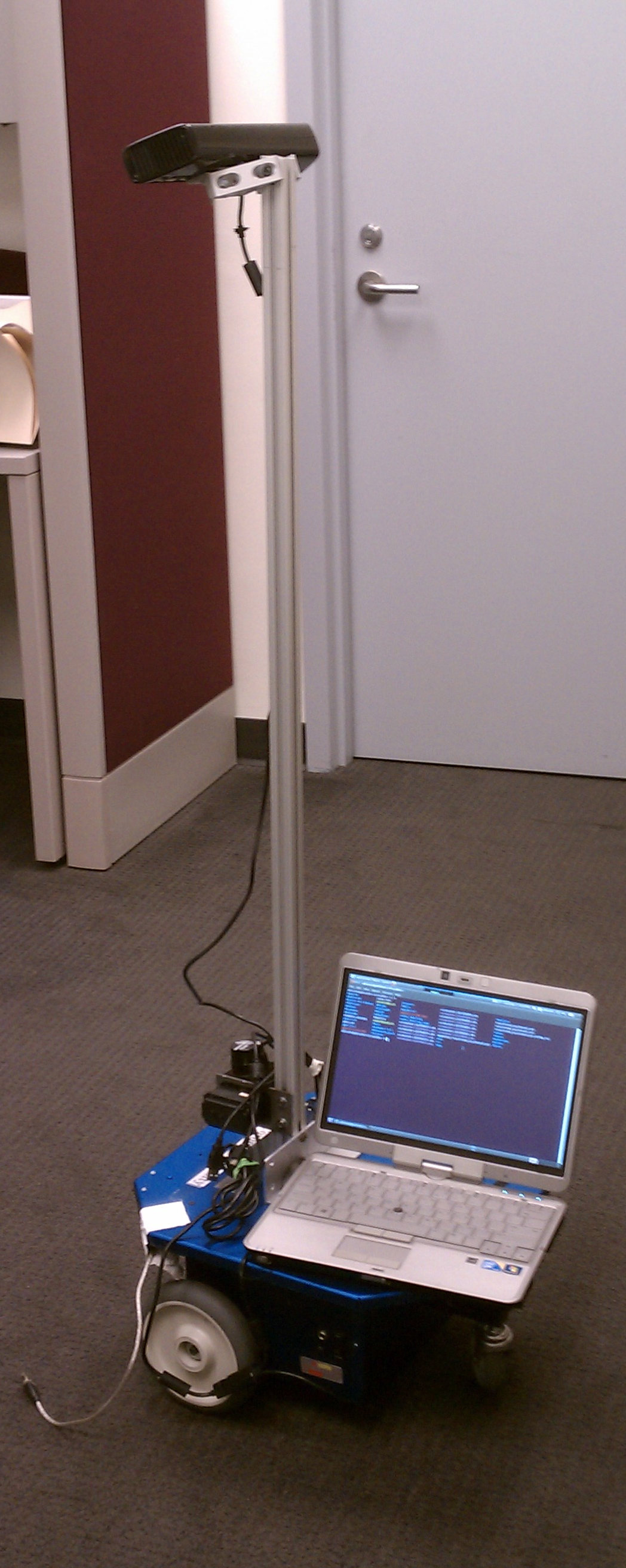}
 \includegraphics[width=0.75\linewidth,height=1.8in]{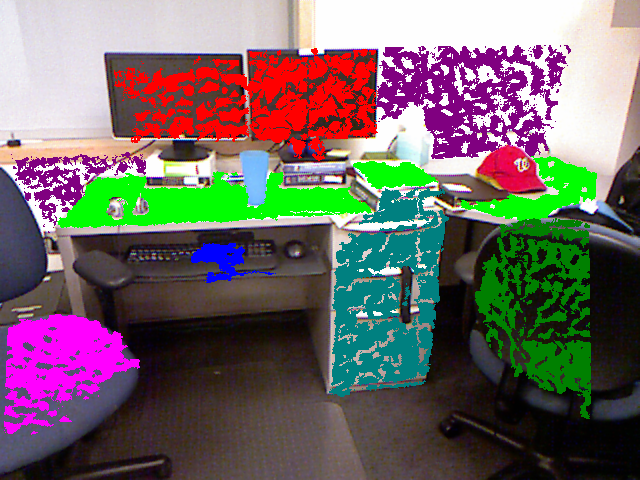}\\
 \includegraphics[width=0.97\linewidth,height=0.2in]{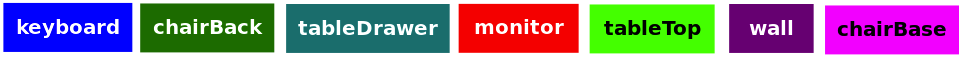}
 \caption{(Left) Cornell's Blue robot mounted with an RGB-D camera (Microsoft Kinect).  (Right) Predicted labeling of a scene. }
\label{fig:robotLabeled}
 \end{figure}

\begin{figure*}[t!]
 \centering
 \includegraphics[width=0.75\linewidth,height=0.25in]{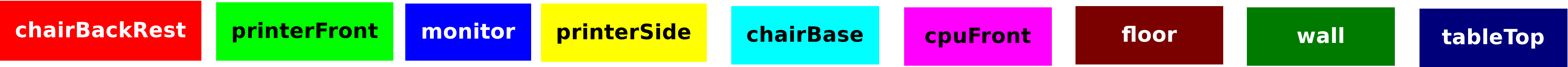}\\
\includegraphics[width=.32\linewidth,height=0.85in]{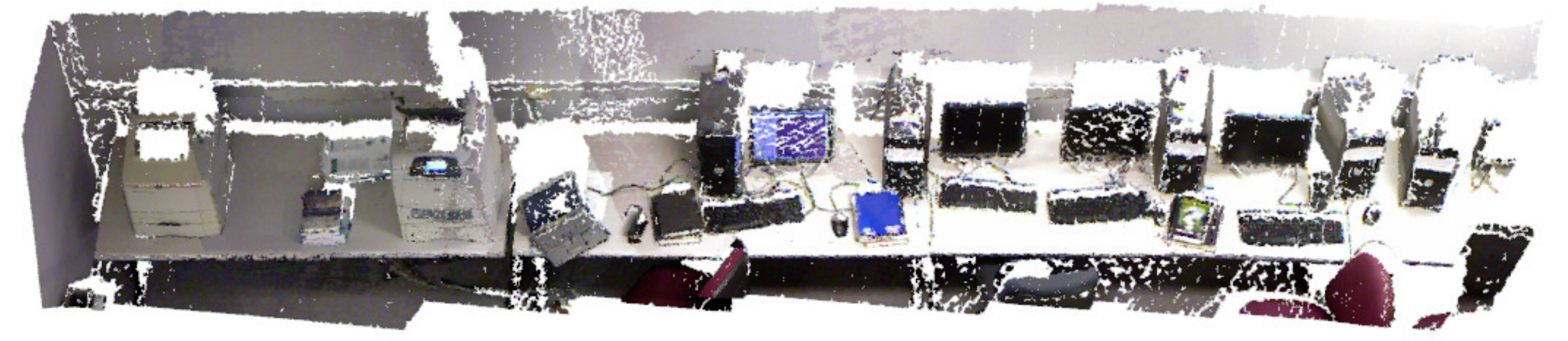} 
\includegraphics[width=.32\linewidth,height=0.85in]{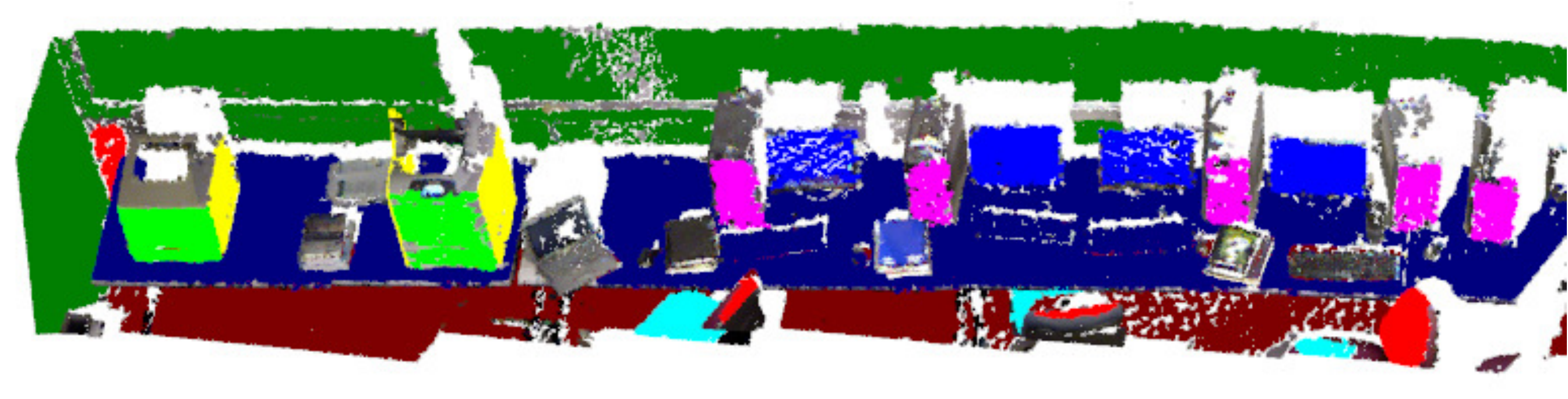} 
\includegraphics[width=.32\linewidth,height=0.85in]{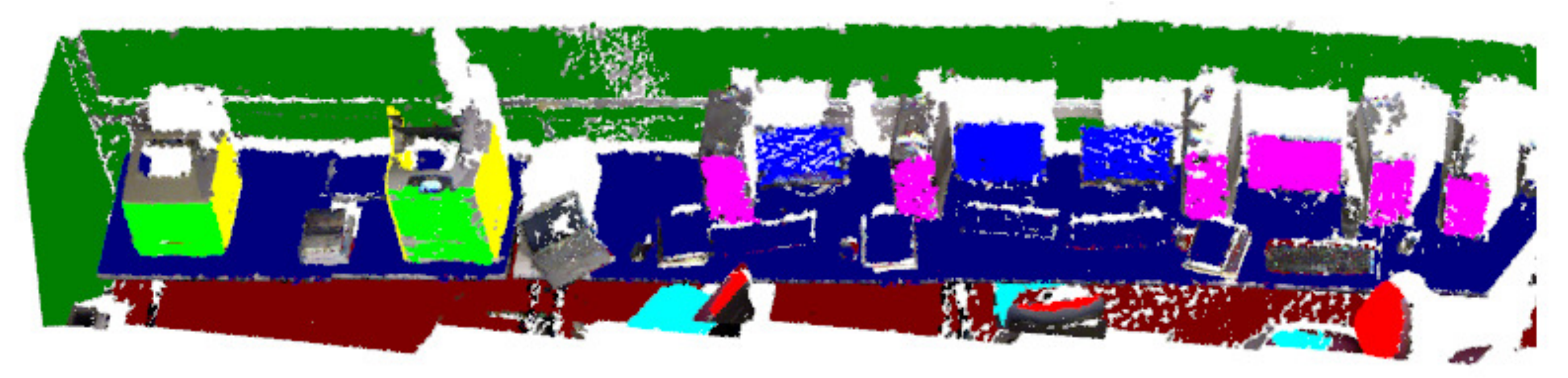} \\
 \includegraphics[width=0.75\linewidth,height=0.25in]{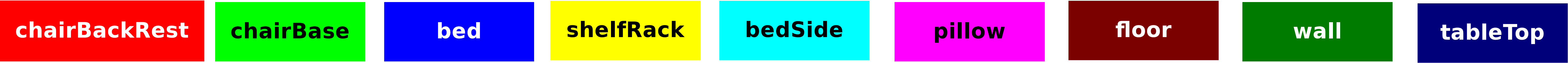}\\
\includegraphics[width=.31\linewidth]{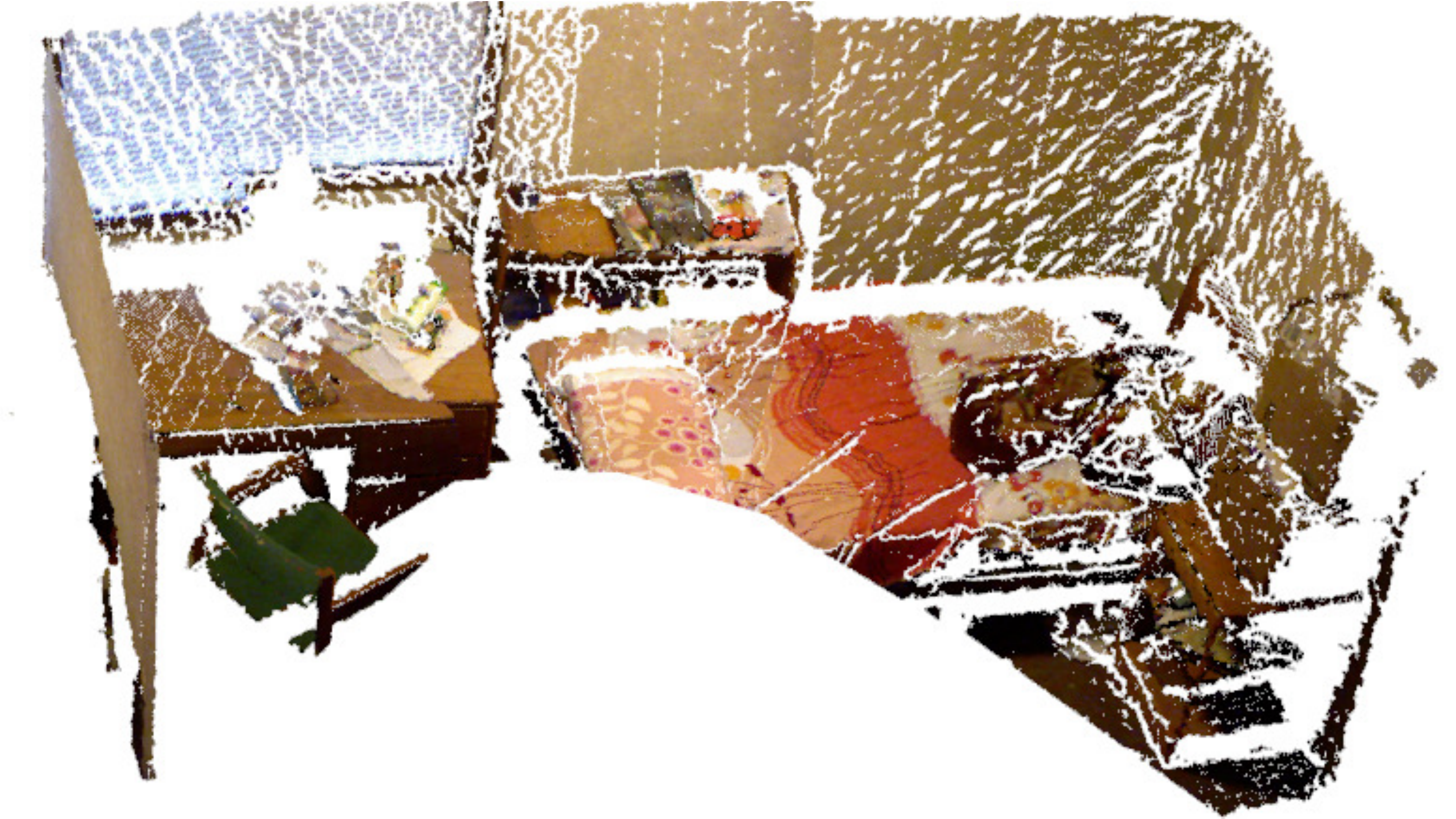} 
\includegraphics[width=.31\linewidth]{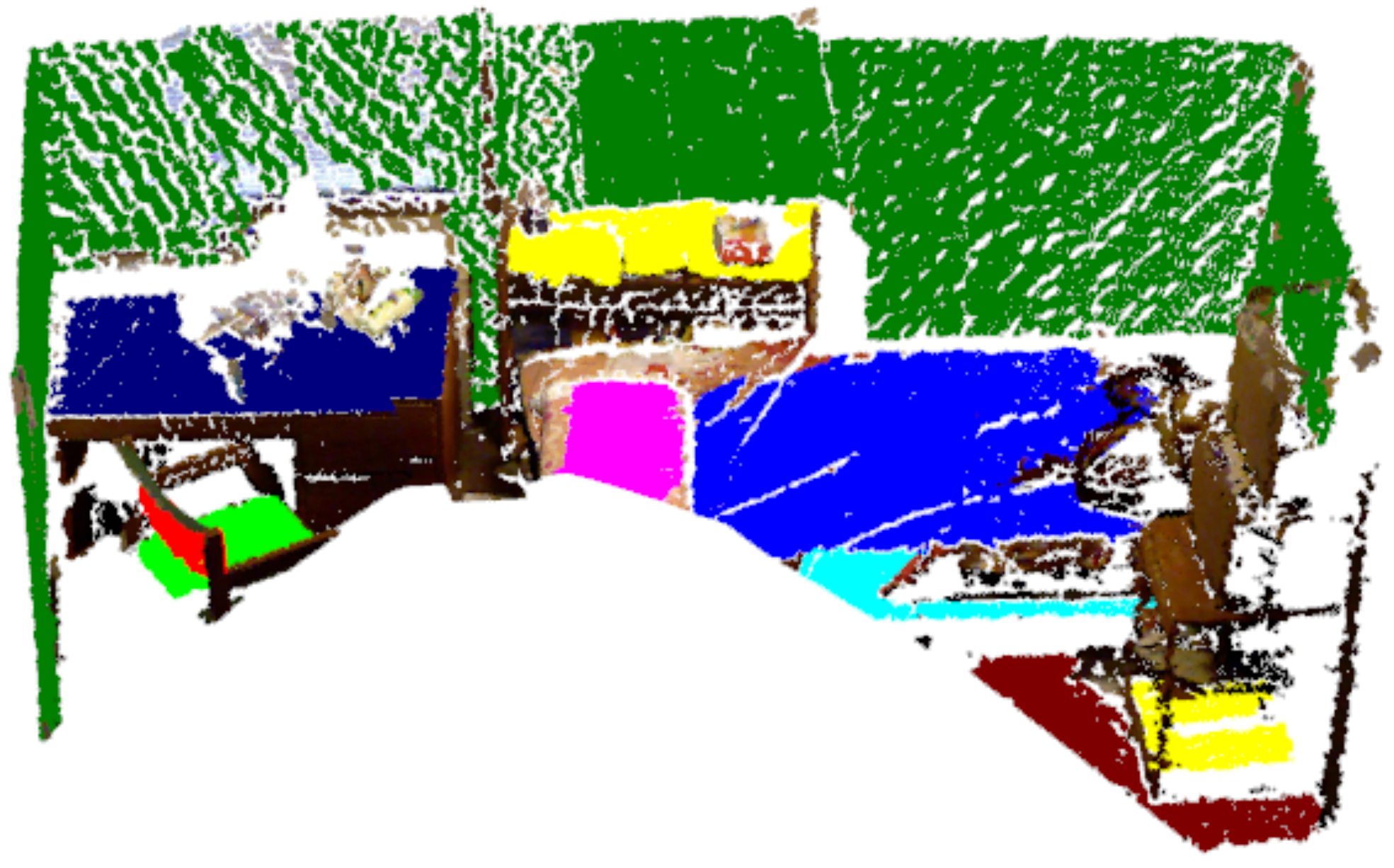} 
\includegraphics[width=.31\linewidth]{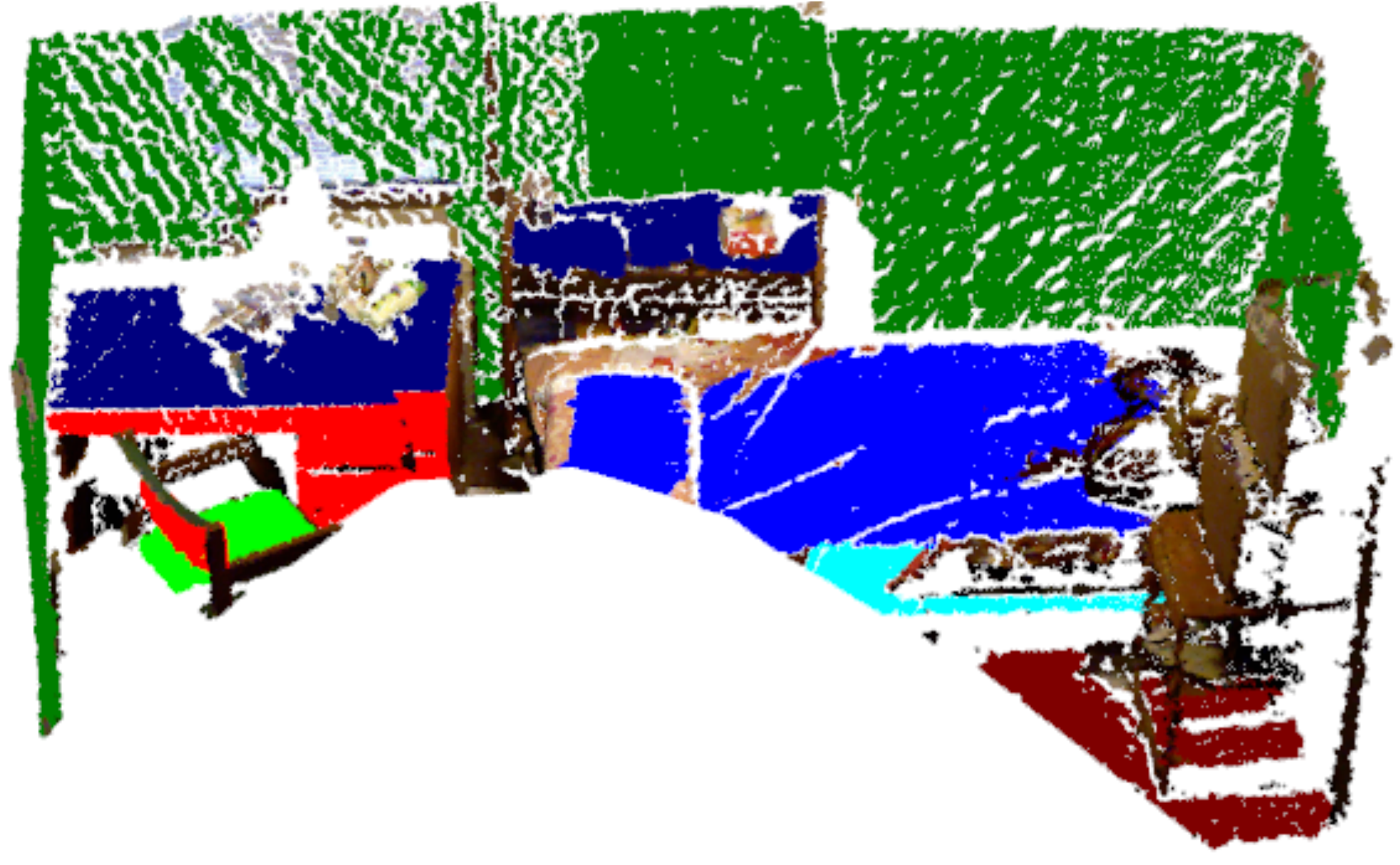} 
 \caption{Office scene (top) and Home (bottom) scene with the corresponding label coloring above the images. The left-most is the original point cloud, the middle is the ground truth labeling and the right most is the point cloud with predicted labels.}
\label{fig:examplePCD}
 \end{figure*}

Some previous works attempt to first infer the 3D structure from 2D images \citep{saxena2005learningdepth,Hoiem:puttingobjects,saxena-make3d-pami,divvala2009empirical} for 
improving object detection. However, the inferred  
3D structure is not accurate enough to give significant improvement.
Another recent work \citep{xiong:indoor} considers labeling a scene using a single 
3D view (i.e., a 2.5D representation).
In our work, we first use SLAM in order to compose multiple views from a 
Microsoft Kinect RGB-D sensor into one 3D point cloud, providing each RGB pixel 
with an absolute 3D location in the scene. We then (over-)segment the scene and predict 
semantic labels for each segment (see Fig.~\ref{fig:examplePCD}). We not only predict coarse 
classes like in \citep{xiong:indoor,Anguelov/etal/05} (i.e., wall, ground, ceiling, building), but also 
label individual objects (e.g., printer, keyboard, monitor). Furthermore, we model rich relational 
information beyond an associative coupling of labels \citep{Anguelov/etal/05}. 

In this paper, we propose and evaluate the first model and learning algorithm for scene 
understanding that exploits rich relational information derived from the full-scene 3D point cloud 
for object labeling and search. In particular, we propose a graphical model that naturally captures the geometric relationships of a 3D scene. Each 3D segment is associated with a node, and pairwise potentials model the  relationships between segments (e.g., co-planarity, convexity, visual similarity, object co-occurrences and proximity). The model admits efficient approximate inference \citep{Kolmogorov/Rother/07}, and we show that it can be trained using a maximum-margin approach \citep{Taskar/AMN,Tsochantaridis/04,Finley/Joachims/08a} that globally minimizes an upper bound on the training loss. We model both associative and non-associative coupling of labels. With a large number of object classes, the model's parsimony becomes important. 
Some features are better indicators of label similarity, while other features
are better indicators of non-associative relations such as geometric arrangement (e.g., \emph{on-top-of}, \emph{in-front-of}). We therefore model them using appropriate clique potentials rather than using general clique potentials.
Our model is thus highly flexible.


We also present an algorithm that uses the contextual information present in a 3D scene 
to predict where an object can be found. For example, its more likely to find a keyboard on top of a table and in front of a 
monitor, and find a table drawer between the table top and the floor.
A robot can use this information in many ways. The robot can move towards the contextually likely location to obtain a better view of the object, resulting in an increase in detection
performance. This is especially helpful for small objects such as a keyboard that often
appear as a segment with only a few points in the original view.
It also helps when an object is not visible in the current view or occluded---the robot can
move to obtain additional views in contextually likely positions of the object.

We extensively evaluate our model and algorithms over a total of 
52 scenes of two types: offices and homes.  These scenes were built from about 550 views 
from the Kinect sensor. We considered the problem of labeling each segment in the scene (from a total of 
about 50 segments per scene) into  27 classes (17 for offices and 
17 for homes, with 7 classes in common). Our experiments show that our method, which captures 
several local cues and contextual properties, achieves an overall performance of 84.06\% on office 
scenes and 73.38\% on home scenes.

We also evaluated our labeling and contextual search algorithms on two mobile 
robots. In particular, in the task of finding 12 objects in 10 cluttered offices, our robot found 
the objects with  96\% precision and 75\% recall.
Fig.~\ref{fig:robotLabeled} shows Cornell's Blue robot which was used in our experiments and a sample output labeling
of an office scene.
We have made the videos, data and the full source code (as a ROS and PCL package)
  available online at: \url{http://pr.cs.cornell.edu/sceneunderstanding}.

 \section{Related Work}
 
There is a huge body of work in the area of scene understanding and object recognition from 2D images. 
Previous works have focussed on several different aspects: designing good local features such as 
HOG (histogram-of-gradients)  \citep{dalal2005histograms}, bag of words \citep{csurka2004visual}, and eigenvectors and eignevalues of the scatter matrix \citep{lalonde:natural}, active vision for robotics \citep[e.g.,][]{jia_roboticobjectdetection}, and designing good global (context) features such as GIST features \citep{GIST}.  
\citeauthor{Collet2011}'s \citeyearpar{Collet2011} MOPED framework performs single-image and multi-image object 
recognition and pose estimation in complex scenes using an algorithm which iteratively estimates 
groups of features that are likely to belong to the same object 
through clustering, and then searches for object hypotheses within each of the groups.

However, the above mentioned approaches
do not consider the relative arrangement of the parts of an object or of different objects with
respect to each other. It has been shown that this contextual information significantly 
improves the performance of image-based object detectors. A number of works
propose models that explicitly capture the relations between different parts of the object, e.g., part-based models \citep{felzenszwalb2008discriminatively}, and between 
different objects in 2D images 
 \citep{Torralba:exploting_context,HeitzECCV_usingstufftofindthings,li2011_thetamrf}. However, a lot of valuable information about the shape and geometric layout of objects is lost when a 2D image is formed from the corresponding 3D world.  In some recent works, 3D layout or depths have been used for improving object detection 
 \citep[e.g.,][]{saxena2005learningdepth,Hoiem:puttingobjects,saxena20083Ddepth,saxena-make3d-pami,hoiem-indoors,hebert-room-layout,Leibe07:dynamic,heitz2008cascaded,li2011feccm,laplaciancrfs_cvpr2012}. 
 Here a rough 3D scene geometry (e.g., main surfaces in the scene)
  is inferred from a single 2D image or a stereo video stream, respectively. 
However, the estimated geometry is not accurate enough to give significant 
improvements.
 With 3D data, we can more precisely determine the shape, size and geometric orientation of the objects, 
 and several other properties and therefore capture much stronger context.

 Visual context can also improve the performance of object detection techniques, by providing cues about an object presence. \citeauthor{Perko2010} \citeyearpar{Perko2010} provided a method to use contextual features and train a classifier which can predict likely locations of objects, also referred to as the `focus of attention', for directing object detection.  In methods using active recognition, the performance of object detection for robotics is improved by letting the robots take
 certain actions, such as moving to an optimal location for obtaining a different view of the object, 
 based on measurement related to entropy \citep[e.g.,][]{Laporte2006,Laporte2004,Denzler2002,Meger2010,Ma2011}. The goal here is to
 obtain high performance in less number of actions.  \citeauthor{ijcai/JiaSC11} \citeyearpar{ijcai/JiaSC11} proposes a path planning method which selects 
 a path where the robot obtains new views of an object, and then these views are used for training the classifiers. 
 However, these methods only use 2D images and do not have advantage of using the rich information 
 present in 3D data. We show that our proposed model captures context which not only helps in labeling the scene but
 also to infer most contextually likely locations of objects using the objects already found in the scene.
 This enables the robot  to move to contextually likely locations and obtain better views for finding objects and improve the performance of scene labeling.   

Some earlier works \cite{gould:fusion,Rusu:ObjectMaps,Dima2004,quigley:high-accuracy} have tried to combine shape and color information from multiple sensors for tasks such as object and obstacle detection. The recent availability of synchronized videos of both color and depth obtained from RGB-D (Kinect-style) 
depth cameras, shifted the focus to making use of both visual as well as 
shape features for  object detection \citep[e.g.,][]{lai:icra11a, lai:icra11b}, 3D segmentation   
\citep[e.g.,][]{Collet:icra2011}, human pose estimation \citep[e.g.,]{ly_humanpose}, and  human activity detection \citep{sung_activity}.
These methods demonstrate that augmenting 
visual features with 
3D information can enhance object detection in cluttered, real-world environments. 
 However, these works do not make use of the contextual 
relationships between various objects which have been shown to be useful for tasks such as object 
detection and scene understanding in 2D images.
Our goal is to perform  semantic labeling of indoor scenes by modeling and learning several contextual
relationships. 



  There is also some recent work in labeling outdoor scenes obtained from LIDAR data into
  a few geometric classes (e.g., ground, building, trees, vegetation, etc.).
  \citeauthor{Golovinskiy:shape-basedrecognition} \citeyearpar{Golovinskiy:shape-basedrecognition} 
 capture context by designing node features.   \citeauthor{xiong:3DSceneAnalysis} \citeyearpar{xiong:3DSceneAnalysis} do so by stacking layers of classifiers;
however this models only limited correlation between the labels.
An Associative Markov Network is used in \citep{Munoz2009:mrf,Anguelov/etal/05,xiao2009multiple} to favor similar labels for nodes in the cliques.
However, many relative features between objects are not associative in nature. 
For example, the relationship \emph{on-top-of} does not hold in between two ground segments, i.e., a ground segment cannot be \emph{on-top-of} another ground segment. Therefore, using an associative Markov network is very restrictive for our problem.  
Contemporary work by \citep{Shapovalov2011} did address this issue by using a cutting plane method to train non-associative Markov network. However, as we later show in our experiments, a fully non-associative Markov Model is not ideal when we have a large number of classes and features. 
More importantly, many of the works discussed above  \citep[e.g.,][]{Shapovalov2010,Shapovalov2011,Munoz2009:mrf,Anguelov/etal/05,xiong:3DSceneAnalysis} were designed for outdoor scenes with LIDAR data (without RGB values).
Using RGB information together with depths presents some new challenges,
such as designing RGB features for 3D segments. 
Also, since we consider much larger number of classes compared to previous works 
(17 vs 3-6 for previous works), the learning task is more challenging due to the large number of parameters. We address this by proposing a parsimonious model.

The most related work to ours is \citep{xiong:indoor}, where they label the planar patches in a point cloud of 
an indoor scene with four geometric labels (walls, floors, ceilings, clutter).  They use a CRF to 
 model geometrical relationships such as orthogonal, parallel, adjacent, and coplanar.
 The learning method for estimating the parameters in \citep{Douillard2011,xiong:indoor} was based on maximizing the pseudo-likelihood
 resulting in a sub-optimal learning algorithm.
In comparison, our basic representation is 3D segments (as compared to planar patches) and 
we consider a much larger number of classes (beyond just the geometric classes).
We also capture a much richer set of relationships between pairs of objects, and
use a principled max-margin learning method to learn the parameters of our model.

\section{Approach}\label{S.properties}

 \begin{figure}[t!]
 \centering

\includegraphics[width=.95\linewidth,height=1.5in]{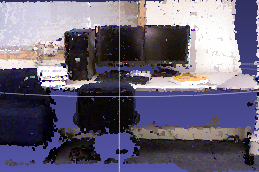} \\
\vskip 0.1in
\includegraphics[width=.95\linewidth,height=1.5in]{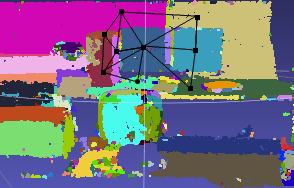} 
 \caption{ (Top) Point cloud of an office scene containing 2 monitors and a CPU on a table and a chair beside it. (Bottom) The segmentation output of the point cloud with each segment 
 represented with a different color.  The black dots and lines represent the nodes and edges of the undirected graph defined over the segments (for clarity not all nodes and edges are shown).   }
\label{fig:segments}
 \end{figure}

We now outline our approach, including the model, its inference methods, and the learning algorithm. 
Our input is multiple Kinect RGB-D images of a scene (i.e., a room) stitched into a single 
3D point cloud using RGBDSLAM.\footnote{http://openslam.org/rgbdslam.html} Each such combined point cloud is 
then over-segmented based on smoothness (i.e., difference in the local surface normals) and 
continuity of surfaces  (i.e.,  distance between the points). Fig.~\ref{fig:segments} shows the 
segmentation output for an office scene. These segments are the 
atomic units in our model. Our goal is to label each one of them.


Before getting into the technical details of the model, we first outline the properties we aim 
to capture in our model below:



\smallskip
\noindent
\textbf{Visual appearance.} The reasonable success of object detection in 2D images 
shows that visual appearance is a good indicator for labeling scenes.
We therefore model the local color, texture, gradients of intensities, etc.~for predicting the 
labels.
In addition, we also model the property that if nearby segments are similar in 
visual appearance, they are more likely to belong to the same object.

\smallskip
\noindent
\textbf{Local shape and geometry.}  Objects have characteristic 
shapes---for example, a table is horizontal, a monitor is vertical, a keyboard is uneven, 
and a sofa is usually smoothly curved. 
Furthermore, parts of an object often form a convex shape. We compute
3D shape features to capture this.

\smallskip
\noindent 
\textbf{Geometrical context.}
Many sets of objects occur in characteristic relative geometric configurations. 
For example, a monitor is always \emph{on-top-of} a table, chairs are usually found \emph{near} tables, a keyboard is \emph{in-front-of} a monitor.
This means that our model needs to capture non-associative relationships 
(e.g., that neighboring segments differ in their labels in specific patterns). 

\smallskip
Note the examples given above are just illustrative. For any particular practical application, 
there will likely be other properties that could also be included. As demonstrated in the 
following section, our model is flexible enough to include a wide range of features.

\begin {comment}
\item \textbf{Visual appearance.}  We obtain the RGB values for the points in the segment. The
color and texture of a segment is often a good indicator of what object it is.  
We used histogram-of-gradients (HOG) features that have been successfully used in many object 
recognition tasks for images \citep{dalal2005histograms,felzenszwalb2008discriminatively}. 
We also compute average hue, saturation and intensity values of a segment and their histograms. 
\item \textbf{Local shape.}  Different objects have different shapes---for example, a table is flat while a keyboard
is uneven. We incorporate several shape properties based on surface normals, and other statistics
of the point cloud in the segment. We describe these in Section~\ref{sec:features}.
\item \textbf{Co-occurence of the objects.}  Objects 
such as monitor and keyboard, or pillow and bed, usually co-occur. We need to capture these 
co-occurrence statistics in our model.
\item \textbf{Geometric properties.}  Certain objects follow strong relative location preferences,
such as ``infront-of" or ``on-top-of" 
For example, a monitor is almost always \emph{on-top-of} 
a table and a keyboard is \emph{infront-of} a monitor.  
We model several geometric relations between a pair of
 segments:  relative depth-ordering (e.g., ``infront-of", ``behind"), relative horizontal-ordering 
 (``above", ``below" and ``same height"), coplanar-ness, etc.   See Section~\ref{sec:relfeatures}
 for more details.
\end{comment}


\subsection{Model Formulation}
\label{sec:model}
We model the three-dimensional structure of a scene using a model isomorphic to a Markov Random Field with log-linear node and pairwise edge potentials. Given a segmented point cloud $\x=(\xs{1},...,\xs{N})$ consisting of segments $\xs{i}$, we aim to predict a labeling $\y=(\ys{1},...,\ys{N})$ for the segments. Each segment label $\ys{i}$ is itself a vector of $K$ binary class labels $\ys{i}=(\ysc{i}{1},...,\ysc{i}{K})$, with each $\ysc{i}{k} \in \{0,1\}$ indicating whether a segment $i$ is a member of class $k$. Note that multiple $\ysc{i}{k}$ can be $1$ for each segment (e.g., a segment can be both a ``chair'' and a ``movable object''). We use such multi-labelings in our attribute experiments where each segment can have multiple attributes, but not in segment labeling experiments where each segment can have only one label. 

For a segmented point cloud $\x$, the prediction $\hat{\y}$ is computed as the argmax of a discriminant function $\df{\x}{\y}{\w}$ that is parameterized by a vector of weights $\w$.
\begin{equation} \label{eq:argmax}
\hat{\y} = \argmax_\y \df{\x}{\y}{\w}
\end{equation}
The discriminant function captures the dependencies between segment labels as defined by an undirected graph $(\mathcal{V},\mathcal{E})$ of vertices $\mathcal{V} = \{1,...,N\}$ and edges $\mathcal{E} \subseteq 
\mathcal{V} \times \mathcal{V}$. This undirected graph is derived from the point cloud by adding a vertex for every 
segment in the point cloud and adding an edge between vertices based on the spatial proximity of the corresponding 
segments. 
In detail, we connect two segments (nodes) $i$ and $j$ by an edge if there exists a point in segment $i$ and 
a point in segment $j$ which are less than \emph{context\_range} distance apart. This captures the
closest distance between two segments (as compared to centroid distance between the segments)---we 
study the effect of context range more in Section~\ref{sec:experiments}.
Fig.~\ref{fig:segments} shows the graph structure induced by a few segments of an office scene.

 Given $(\mathcal{V},\mathcal{E})$, we define the following discriminant function based on individual segment features $\fn{i}$ and edge features $\fe{t}{i}{j}$ as further described below.
\begin{eqnarray} \label{eq:model}
\df{\x}{\y}{\w} = \sum_{i \in \mathcal{V}} \sum_{k=1}^{K} \ysc{i}{k} \left[\wn{k} \cdot \fn{i} \right] \nonumber \\
+  \sum_{(i,j)\in \mathcal{E}}   \sum_{T_t \in {\cal T}}  \sum_{(l,k)\in T_t} \ysc{i}{l} \ysc{j}{k}  \left[\we{t}{l}{k} \cdot \fe{t}{i}{j}\right] 
\end{eqnarray}

The node feature map $\fn{i}$ describes segment $i$ through a vector of features, and there is one weight vector for each of the $K$ classes. Examples of such features are the ones capturing local visual appearance, shape and geometry. The edge feature maps $\fe{t}{i}{j}$ describe the relationship between segments $i$ and $j$. Examples of edge features are the ones capturing similarity in visual appearance and geometric 
context.\footnote{Even though it is not represented in the notation, note that both the node feature map $\fn{i}$ and the edge feature maps $\fe{t}{i}{j}$ can compute their features based on the full $\x$, not just $\xs{i}$ and $\xs{j}$.}
There may be multiple types $t$ of edge feature maps $\fe{t}{i}{j}$, and each type has a graph over the $K$ classes with edges $T_t$. If $T_t$ contains an edge between classes $l$ and $k$, then this feature map and a weight vector $\we{t}{l}{k}$ is used to model the dependencies between classes $l$ and $k$. If the edge is not present in $T_t$, then $\fe{t}{i}{j}$ is not used.

We say that a type $t$ of edge features is modeled by an associative edge potential if 
the corresponding graph only has self loops, i.e. ${T_t}=\{(k,k)| \forall k=1..K\}$.
And it is modeled by an non-associative edge potential if the corresponding graph is a fully connected graph, i.e. $T_t=\{(l,k)| \forall l,k=1..K\}$. 

 \begin{figure} 
 \includegraphics[width=1\linewidth]{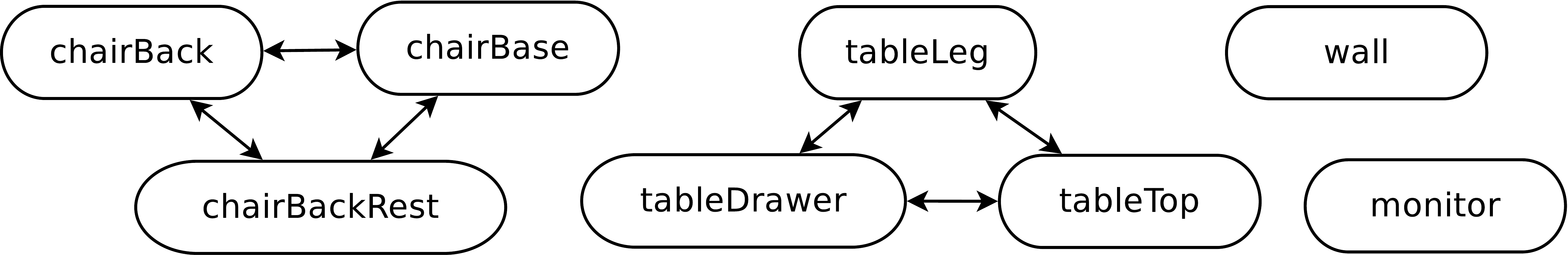}
 \caption{Dependency graph for object-associative features. It also contains self-loops (which we do not show here for clarity).}
 \label{fig:objAssocGraph}
 \end{figure}
 
\noindent
\textbf{Parsimonious model.} 
Most of the works on 2D images only used associative edge potentials 
(e.g., \citep{szeliski2008comparative}), where the idea is that visual similarity is usually an 
indicator of whether two segments (or pixels) belong to same class. 
As discussed before, using only associative edge potentials is very restrictive for the task of object labeling 
and by using non-associative edge potentials we are able to model the dependencies between different objects.  
Examples of such dependencies are geometric arrangements such as \emph{on-top-of};
we usually find monitor, keyboard etc. \emph{on-top-of} a table, rather than a table \emph{on-top-of} a table. 
However, modeling every edge feature via a non-associative edge potential will need $K^2$ parameters  
per edge feature. Therefore, the number of parameters becomes very large with increase in the number of classes and the 
number of edge features. 
Even though,  given sufficient data, a non-associative clique potential is general enough to learn associative 
relationships, this generality comes at an increased cost of training time and memory requirements when the 
number of classes is large.

We make the observation that not all features need to be modeled using non-associative edge potentials that
 relate every pair of classes. 
As described above, certain features such as the visual similarity features indicate when the two segments belong to 
the same class or parts of the same object, where as other geometric features capture relations between 
any pair of classes.  A key reason for distinguishing between object-associative and non-associative features is 
parsimony of the model. As not all edge features are non-associative, we avoid learning weight vectors for 
relationships which do not exist.



Based on this observation we propose our parsimonious model (referred to as \emph{svm\_mrf\_parsimon}) which partitions edge features into two types---object-associative features ($T_{oa}$), such as visual similarity, coplanarity and convexity, which usually indicate that two segments belong to the same object, and
 non-associative features ($T_{na}$), such as relative geometric arrangement (e.g., \emph{on-top-of}, \emph{in-front-of}), which can capture characteristic configurations under which pairs of different objects occur.
We model the object-associative features using object-associative potentials which capture only the dependencies between parts of the same object. The graph for this type of features contains only edges between parts of same object and self loops as shown in Fig.~\ref{fig:objAssocGraph}. Formally, $T_{oa}=\{(l,k) | \exists object , ~ l,k\in {\rm parts}({\rm object})\}$. The non-associative features are modeled as non-associative edge potentials.   Note that $|T_{na}| >> |T_{oa}|$ since, in practice, the number of parts of an 
objects is much less than K. Due to this, the model we learn with both type of edge features will have much lesser 
number of parameters compared to a model learnt with all edge features as non-associative features. We show the 
performance of modeling the edge features using  various types of edge potentials in Section~\ref{sec:experiments}.

\subsection{\label{sec:features}Features}

\begin{figure*}[th!]
\centering
\includegraphics[width=.25\linewidth]{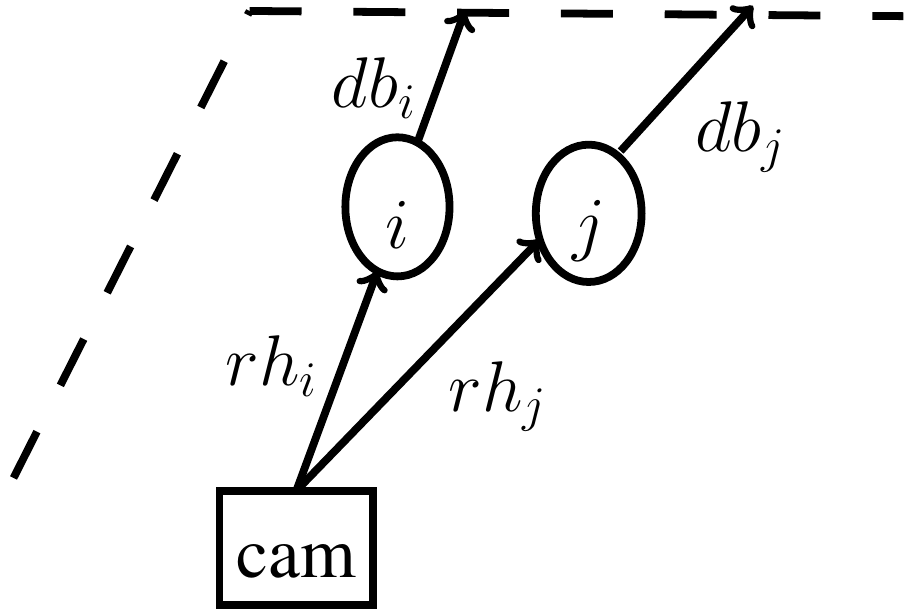}
\hskip 0.3in
\includegraphics[width=.25\linewidth]{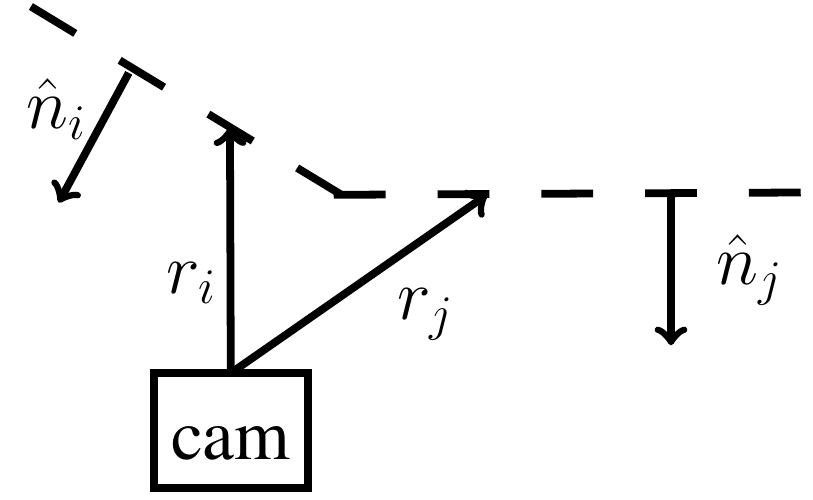}
\hskip 0.3in
\includegraphics[width=.3\linewidth]{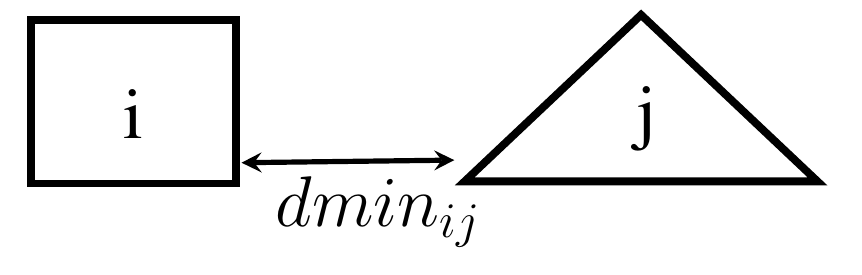}
\caption{{\small Illustration of a few features. (Left) Features N9 and E9. Segment $i$ is \emph{in-front-of} segment $j$ if ${\bf rh_i }<{\bf rh_j}$. (Middle) Two connected segment $i$ and $j$ form a convex shape if  ${\bf (r_i-r_j).\hat{n}_i \ge 0}$ and  ${\bf(r_j-r_i).\hat{n}_j \ge 0}$. (Right) Illustrating feature E8.} }
\label{fig:features}
\end{figure*}

The various properties of objects and relations between them are captured in our model
with the help of various features that we compute from the segmented point clouds. 
Our model uses
two types of features -- the node features and the edge features. 
Since, 
the robot knows the height and orientation of its Kinect sensor, we align 
  all the point clouds 
  so that the z-axis is vertical and the ground is at zero height for computing the features.
  Our features are insensitive to horizontal translation and 
rotation of the camera, but they place a lot of emphasis on the vertical direction  
because gravity influences the shape and relative positions of objects to a large extent.
 Note that unlike images, there is no ambiguity of scale in the input point could and all distance measurements are in meters.

\begin{table}[t!]
\caption{Node and edge features.}
{
\begin{minipage}[b]{8cm}\centering

\begin{tabular}{|p{6.0cm}|c|}
\multicolumn {2}{c}{Node features for segment $i$.} \\ 
\hline 
 Description  & Count\\ \hline
{\bf Visual Appearance} & {\bf 48} \\
\hline
N1. Histogram of HSV color values & 14 \\
N2. Average HSV color values & 3 \\
N3. Average of HOG features of the blocks in image spanned by the points of a segment & \multirow{2}{*}{31} \\
\hline
{\bf Local Shape and Geometry} & {\bf 8} \\
\hline
N4. linearness (${\lambda_{i0}}$ - ${\lambda_{i1}}$), planarness (${\lambda_{i1}}$ - ${\lambda_{i2}}$) & 2 \\
N5. Scatter: ${\lambda_i}_0$ & 1 \\ 
N6. Vertical component of the normal: ${\hat{n_i}}_z$ & 1 \\ 
N7. Vertical position of centroid: ${c_i}_z$ & 1 \\ 
N8. Vert.~and Hor.~extent of bounding box  & 2 \\ 
N9. Dist.~from the scene boundary (Fig.~\ref{fig:features})& 1 \\
\hline
\end{tabular}
\end{minipage} 
\begin{minipage}[b]{8cm}
\vspace{0.5cm}
\centering
\begin{tabular}{|p{6cm}|c|}
\multicolumn {2}{c}{Edge features for edge (segment $i$, segment $j$).}\\ \hline
 Description  & Count\\ \hline
{\bf Visual Appearance (associative)} & {\bf 3} \\
\hline
E1. Difference of avg HSV color values & 3\\
\hline
{\bf Local Shape and Geometry (associative)} & {\bf 2} \\
E2. Coplanarity and convexity (Fig.~\ref{fig:features})
& 2\\ 
\hline
{\bf Geometric context (non-associative)} & {\bf 6} \\
\hline

E3. Horizontal distance b/w centroids.
& 1\\ 

E4. Vertical Displacement b/w centroids:
 $(c_{iz} - c_{jz})$ 
& 1\\ 

E5. Angle between normals (Dot product):
$ \hat{n}_i \cdot \hat{n}_j $
& 1\\ 

E6. Difference in angle with vertical: & \multirow{2}{*}{1} \\
$ \cos^{-1}({n_{iz}})$ - $\cos^{-1}({n_{jz}}) $
& \\

E8. Distance between closest points:
$ \min_{u\in s_i,v\in s_j} d(u,v) $ (Fig.~\ref{fig:features})
& \multirow{2}{*}{1}\\
 
E9. Relative position from camera (\emph{in-front-of}/\emph{behind}).
(Fig.~\ref{fig:features})
& \multirow{2}{*}{1}\\ 

\hline

\end{tabular}
\end{minipage}
}
\label{tab:Features}
\end{table}

Table \ref{tab:Features}  summarizes the features used in our experiments. 
${\lambda_i}_0,{\lambda_i}_1$ and ${\lambda_i}_2$ are the first three eigenvalues  of the scatter matrix computed from the points of segment $i$.  $s_i$ is the set of points in segment $i$ and $c_i$ is the centroid of  segment $i$. 
$r_i$ is the ray vector to the centroid of segment $i$ from the camera position when it was captured. $rh_i$ is the projection of $r_i$ on horizontal plane. 
$\hat{n}_i$ is the unit normal of segment $i$ which points towards the camera ($r_i . \hat{n}_i <0$).
  The node features $\phi_n(i)$ (Table \ref{tab:Features}-top) 
consist of visual appearance features based on histogram of HSV values and the histogram of gradients
(HOG) \citep{dalal2005histograms}, as well as local shape and geometry features that capture properties such 
as how planar a segment is, its absolute location above ground, and its shape.
The local shape features commonly used in the spectral analysis of point clouds (N4-N5) are given by  ${\lambda_i}_2 - {\lambda_i}_1$ (linearness),  ${\lambda_i}_2 - {\lambda_i}_1$ (planarness) and ${\lambda_i}_0$ (scatterness).
N9 captures the tendency of some objects to be near the scene boundaries (walls), e.g., shelf, table, etc.

The edge features $\phi_t(i,j)$ (Table \ref{tab:Features}-bottom) consist of associative features (E1-E2)
based on visual appearance and local shape, as well as non-associative features (E3-E9) that capture
the tendencies of two objects to occur in certain configurations.
 The local shape features include coplanarity and convexity. Coplanarity is defined as:\\
\[
{\small
Coplanarity(s_i,s_j) = \left\{
  \begin{array}{l l}
    -1 & \quad  abs(\hat{n}_i \cdot \hat{n}_j) \le \cos{\alpha}  \\
    1/d_{ij} & \quad \mbox{otherwise} \\
  \end{array} \right. 
}
\] 
where $d_{ij} = abs(( r_i - r_j) \cdot  \hat{n_i}  )$ is the distance between centroids in the direction of the normal.  Coplanarity only makes sense if the planes are almost parallel. So we use a tolerance angle $\alpha$ (which was set to 30 degrees). The value of coplanarity feature is inversely proportional to the distance between the segments when they are parallel, a large value when the segments are coplanar and -1 when they are not parallel.
Convexity determines if the two adjoined segments form a convex surface and is defined as: \\
\[
{\small
Convexity(s_i,s_j) = \left\{
  \begin{array}{l l}
    1 & \mbox  (dmin_{ij} < \tau) \wedge [ [ ( \hat{n}_i \cdot   \overrightarrow{d_{ij}} \le 0 ) \\
       &  \wedge   ( \hat{n}_j \cdot   \overrightarrow{d_{ji}} \le 0) ] \vee \hat{n}_i \cdot \hat{n}_j \le \cos{ \alpha} ]  \\ 
    0 &  \mbox{otherwise}\\
  \end{array} \right.
 }
 \]
where  $dmin_{ij}$ is the minimum distance between segments $s_i$ and $s_j$, $\overrightarrow{d_{ij}} = (r_j  - r_i )$, is the displacement vector from  $r_i$  to  $r_j$  and 
 $\overrightarrow{d_{ji}} =  - \overrightarrow{d_{ij}} $.  $\tau$ and $\alpha$ are the tolerance values for minimum distance and the angle respectively.
The geometric configuration features include features that capture relationships such as \emph{on-top-of} 
and \emph{in-front-of}. These are encoded by the vertical distance between centroids (E4), which is positive 
if segment $i$ is \emph{on-top-of} segment $j$ and negative otherwise, and the relative distance from the camera (E9),
which is positive if segment $i$ is \emph{in-front-of} segment $j$ and negative otherwise.

Finally, all the features are binned using a cumulative binning strategy. 
Each feature instance is represented by $n$ binary values (each corresponding to a bin), where $n$ is the number of bins.
$n$ thersholds are computed where $i^{th}$ one is the $\frac{100i}{n}^{th}$ percentile of values of that features in the dataset.
The value corresponding to the $i^{th}$
bin is set to 1 if the feature value for that instance lies in the range $[min, th_i)$, 
where $th_i$ is the $i^{th}$ threshold.
This gives us a new set of binary features which are then used for learning the model 
and during inference. 
Binning helps in capturing various non-linear functions of features and hence, significantly improves prediction accuracies. In our experiments we use 10 bins for every non-binary feature.

\begin {comment} 
\vspace*{\subsectionReduceTop}
\subsection{Relative Features}
\vspace*{\subsectionReduceBot}

\todo{This section describes the relative features in more detail.}

We model the following geometric relations between a pair of
 segments:  relative depth-ordering, relative horizontal-ordering, coplanar-ness, etc.  
For any given pair of segments, the relative depth-ordering  captures the ``infront-of", ``behind" and 
``none" relations and the relative  horizontal-ordering captures the ``above", ``below" and ``same height" relations.
\end{comment}


\subsection{Computing Predictions\label{sec:computingPredictions}}
Our goal is to label each segment in the segmented point cloud with the most appropriate semantic label. 
This is achieved by finding the label assignment which maximizes the 
value of the discriminant function in Eq.~(\ref{eq:model}). Given the learned parameters of the model and 
the features computed from the segmented point cloud, we need to solve the argmax in Eq.~(\ref{eq:argmax}) for the discriminant function in Eq.~(\ref{eq:model}). This is NP hard.
It can be equivalently formulated as the following mixed-integer program.  
%
%
%

\begin{eqnarray}
\hat{\y}=\argmax_{\y}\max_{\mathbf z} \sum_{i \in \mathcal{V}} \sum_{k=1}^{K} \ysc{i}{k} \left[\wn{k} \cdot \fn{i} \right] \nonumber\\
+\!\!\!\sum_{(i,j)\in \mathcal{E}}  \sum_{T_t \in {\cal T}} \sum_{(l,k)\in T_t} \zsc{ij}{lk} \left[\we{t}{l}{k} \cdot \fe{t}{i}{j}\right] 
 \label{eq:relaxobj}\\
 \forall i,j,l,k: \:\: \zsc{ij}{lk}\le \ysc{i}{l}, 
\zsc{ij}{lk}\le \ysc{j}{k},\:\:\:\: \ysc{i}{l} + \ysc{j}{k} \le \zsc{ij}{lk}+1 \nonumber \\
\zsc{ij}{lk},\ysc{i}{l} \in \{ 0,1 \} \label{eq:relaxconst}\\
\forall i: \sum_{j=1}^{K} \ysc{i}{j} = 1 \label{eq:relaxconstcost}
\end{eqnarray}
Note that the products $\ysc{i}{l} \ysc{j}{k}$ have been replaced by auxiliary variables $z^{lk}_{ij}$. We can  compute the exact mixed integer solution using a general-purpose MIP solver.\footnote{http://www.gnu.org/software/glpk}
 We use this method for inference
in our offline object labeling experiments (described in Section~\ref{sec:labelingresults}) and set a time limit of 30 minutes for the MIP solver. This takes 18 minutes on average for a full-scene point cloud and 2 minutes on average for a single-view point cloud. All runtimes are for single CPU implementations using 17 classes.

However, when using our algorithm on labeling new scenes (e.g., during our robotic experiments), objects other than the 27 objects we modeled might be present.
Forcing the model to predict one of the 27 objects 
for all segments would result in wrong predictions for every segment of an unseen class. Therefore,  we relax the constraints $\forall i: \sum_{j=1}^{K} \ysc{i}{j} = 1$ in Eq.~(\ref{eq:relaxconstcost}) to 
$\forall i: \sum_{j=1}^{K} \ysc{i}{j} \le 1$, which allows 
a segment to remain unlabeled. This increases precision 
significantly at the cost of some drop in recall. 
Also, this relaxed MIP 
only takes 30 seconds 
on an average for a single-view point cloud.

We further relax the problem by removing the constraints in 
Eq.~(\ref{eq:relaxconstcost}) and let the variables $\zsc{ij}{lk}$ and $\ysc{i}{l}$ take values in the interval $[0,1]$. This results in a linear program that can be shown to always have half-integral solutions (i.e., $\ysc{i}{l}$ only take values $\{0,0.5,1\}$ at the solution) \citep{hammer1984roof}.  Furthermore, this relaxation can also be solved as a QPBO (Quadratic Pseudo-Boolean Optimization) problem using a graph-cut method\footnote{http://www.cs.ucl.ac.uk/staff/V.Kolmogorov/software/QPBO-v1.3.src.tar.gz} \citep{Kolmogorov/Rother/07}, which is orders of magnitude faster than using a general purpose LP solver. 
The graph-cut method takes less than \textit{\textbf{0.05 seconds}} for labeling any 
full-scene or single-view point cloud.
We refer to the solution of this relaxation as $\hat{\y}_{cut}$.

The relaxation solution $\hat{\y}_{cut}$ has an interesting property called {\em Persistence} \citep{Boros/Hammer/02,hammer1984roof}. Persistence says that any segment for which the value of $\ysc{i}{l}$ is integral in $\hat{\y}_{cut}$ (i.e., does not take value $0.5$) is labeled just like it would be in the optimal mixed-integer solution. Note that in all relaxations, we say that the node $i$ is predicted 
as label $l$ iff $\ysc{i}{l} =1$.

In Section \ref{sec:experiments}, Table~\ref{tbl:detection_results} shows that the solution $\hat{\y}_{cut}$, despite being a relaxation, achieves similar results on all metrics for both full-scene and single-view point clouds, and is orders of magnitude faster to compute.

\subsection{Learning Algorithm}

We take a large-margin approach to learning the parameter vector $\w$ of Eq.~(\ref{eq:model}) from labeled training examples $(\x_1,\y_1),...,(\x_\n,\y_\n)$ \citep{Taskar/MMMN,Taskar/AMN,Tsochantaridis/04}. Compared to Conditional Random Field training \citep{Lafferty/etal/01} using maximum likelihood, this has the advantage that the partition function 
need not be computed, and that the training problem can be formulated as a convex program for which efficient algorithms exist.

Our method optimizes a regularized upper bound on the training error
\begin{equation}\label{eq:emprisk}
R(h) = \frac{1}{\n} \sum_{j=1}^{\n} \loss{\y_j}{\hat{\y}_j},
\end{equation}
where  $\hat{\y}_j$ is the optimal solution of Eq.~(\ref{eq:argmax}), $\loss{\y}{\hat{\y}}=\sum_{i=1}^{N} \sum_{k=1}^{K} |\ysc{i}{k} - \hat{\ysc{i}{k}}|$, and $h$ is the function mapping the input $\x$ to an output $\y$. In our training problem, the function $h$ parameterized by $w$ is $h_w(\x) = \hat{\y}$. To simplify notation, note that Eq.~(\ref{eq:relaxobj}) can be equivalently written as $\w^T \Psi(\x,\y)$ by appropriately stacking the $\wn{k}$ and $\we{t}{l}{k}$ into $\w$ and the $\ysc{i}{k}\fn{k}$ and $\zsc{ij}{lk}\fe{t}{l}{k}$ into $\Psi(\x,\y)$, where each $\zsc{ij}{lk}$ is consistent with Eq.~(\ref{eq:relaxconst}) given $\y$. Training can then be formulated as the following convex quadratic program \citep{joachims2009cutting}:
\begin{eqnarray} \label{eq:trainqp}
&  \min_{w,\xi}    & \frac{1}{2} \w^T\w + C\xi\\
&  s.t.\   & \forall \bar{\y}_1,...,\bar{\y}_\n \in \{0,0.5,1\}^{N \cdot K} :  \cr
& & \frac{1}{n} \w^T \sum_{i=1}^{n} [\Psi( \x_i, \y_i) \nonumber - \Psi(\x_i,\bar{\y}_i)] \ge \Delta(\y_i,\bar{\y}_i) -\xi \nonumber
\end{eqnarray}
While the number of constraints in this quadratic program is exponential in $\n$, $N$, and $K$, it can nevertheless be solved efficiently using the cutting-plane algorithm for training structural SVMs \citep{joachims2009cutting}. The algorithm maintains a working set of constraints, and it can be shown to provide an $\epsilon$-accurate solution after adding at most $O(R^2 C / \epsilon)$ constraints (ignoring log terms). The algorithm merely needs access to an efficient method for computing
\begin{eqnarray}
\bar{\y}_i & = & \!\!\!\!\!\argmax_{\y \in \{0,0.5,1\}^{N \cdot K}} \left[ \w^T \Psi(\x_i,\y) + \loss{\y_i}{\y} \right].
\end{eqnarray}
Due to the structure of $\loss{.}{.}$, this problem is identical to the relaxed prediction problem in Eqs.~(\ref{eq:relaxobj})-(\ref{eq:relaxconst}) and can be solved efficiently using graph cuts.

Since our training problem is an overgenerating formulation as defined in \citep{Finley/Joachims/08a}, the value of $\xi$ at the solution is an upper bound on the training error in Eq.~(\ref{eq:emprisk}). Furthermore, \citeauthor{Finley/Joachims/08a} \citeyearpar{Finley/Joachims/08a} observed empirically that the relaxed prediction $\hat{\y}_{cut}$ after training $\w$ via Eq.~(\ref{eq:trainqp}) is typically largely integral, meaning that most labels $\ysc{i}{k}$ of the relaxed solution are the same as the optimal mixed-integer solution due to persistence.  We made the same observation in our experiments as well:
specifically, the average percentage of variables per example that are labeled with integral values is 98.54\%.

\section{Contextual Search}
\label{sec:contextualSearch}

In robotic tasks such as of finding objects, a robot might not find the object it was 
looking for in its current view. 
The object might be far away,  or  occluded, or even out of view. 
In this section, we propose a method which determines the optimal location for the robot to move 
for finding the object it was looking for. Our method
uses the context from the objects already identified in the robot's current view.

Formally, the goal is to find the 3D location where the desired object is most likely to be found, given a (partially) labeled point cloud. 
This can be done by sampling 3D locations in the current view and using the learned discriminant function to compare the chances of finding the desired object at those locations. 
We generate these samples by discretizing the 3D bounding box of the labeled point cloud and considering the centre of each grid as a sample. We considered 1000 equally spaced samples in our
experiments.

 \begin{figure}[htb!]
 \centering
 \includegraphics[ width=0.45\textwidth]{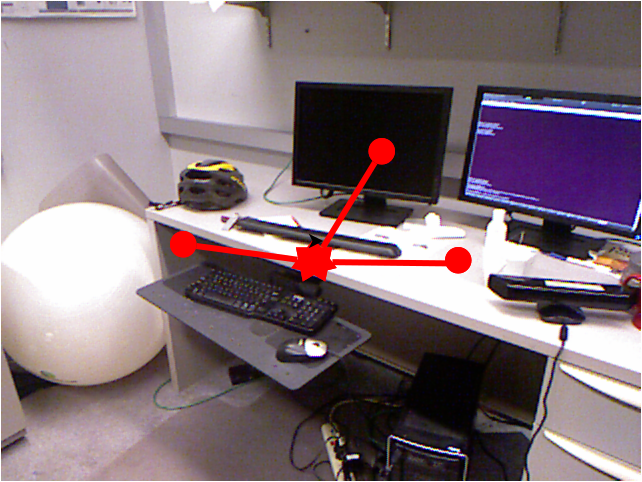} 
 \caption{Figure illustrating our method for contextual search. The location of the hallucinated segment is shown by a red star and its neighbors are denoted using red edges. In this particular scene the hallucinated keyboard segment is connected to monitor, table-top and table-leg segments. The keyboard and tray segments were left unlabeled by the algorithm, so they are not connected.}
\label{fig:contextSearch}
\end{figure}

\subsection{Augmented Graph}

To evaluate the score of a sample ($l_x,l_y,l_z$) 
  we first generate a graph on which the discriminant function can be applied.
For this, we first take the graph of labeled segments (labeled using our algorithm) and augment it with a node $h$ corresponding to a hallucinated segment $x_h$ at location ($l_x,l_y,l_z$). 
We then add edges 
 between this node, $h$, and the nearby nodes (labeled segments) that
 are within $context\_range$ distance from the sampled location. We denote these neighboring segments of the hallucinated segment by $Nbr(x_h)$. Suppose we are looking for an (missing) object class $k$, we label the newly added node as class $k$, i.e., the $k^{th}$ element of $y_h$ is 1 and rest are 0.
Fig.~\ref{fig:contextSearch} illustrates one such augmentation step.  
We can now apply our discriminant function to this augmented graph.
Note that each of our sampled locations will give us a unique corresponding augmented graph.



\subsection{Discriminant Function and Inference}

We use the discriminant function from Eq.~(\ref{eq:model}) to compute the score of each augmented graph. 
The optimal sample (location) for finding an instance of class $k$, $OL(k)$, is the one with highest value of the 
discriminant function applied to its corresponding augmented graph.
Note that only the location-dependent terms in the discriminant function for the newly added node and edges
can effect its value, since the rest of the terms are same for every sample. 
Thus, we can compute $OL(k)$ very efficiently.

We denote the sum of all terms of the discriminant  function which do not depend 
on the location $(l_x,l_y,l_z)$ by a constant $\rho$.
We denote to the label of a node $j$ in the original labeled graph (before augmentation) as $label(j)$.
Let $\fnp{h}$ denote the features of the hallucinated node $h$, which only depend on its location ($l_x,l_y,l_z$). Similarly, let $\fep{t}{h}{j}$ denote the features of the edge $(h,j)$ which only depend on location of the hallucinated node.
Let $w_n'$ and $w_t'$ denote the projection of node and edge weight vectors respectively that
 contain only the components
 corresponding to the location-dependent features.
 Now we can formally define the optimal location as:

\begin{equation}
 OL(k) = \argmax_{x_h \in samples} \df{\x \cup x_h}{\y \cup y_h }{\w} \\
  \label{eq:contextsearch}
\end{equation}
where,
\begin{eqnarray*}
\df{\x \cup x_h}{\y \cup y_h }{\w} =  \wnp{k} \cdot \fnp{h}  \\
+  \sum_{(j)\in Nbr(x_h)} \sum_{T_t \in {\cal T}}  
\left[ \wep{t}{k}{label(j)} \cdot \fep{t}{h}{j} \right. \\
\left.  + \wep{t}{label(j)}{k} \cdot \fep{t}{j}{h} \right] 
+ \rho
\end{eqnarray*}

Once the optimal location $OL(k)$ is found by solving Eq.~(\ref{eq:contextsearch}), the robot moves close to this location  to find objects as described in our contextually-guided object detection experiments presented in Section \ref{sec:robotExps}.

   \section{Experiments\label{sec:experiments}}

 \begin{figure} 
 \centering
 \includegraphics[height=0.45\textheight, width=0.3\textwidth]{blue_robot.jpg} 
 \caption{ Cornell's Blue Robot. It consists of a mobile base mounted with a Kinect sensor.}
\label{fig:blueRobot}
\vskip 0.6in
 \centering
 \includegraphics[height=0.45\textheight,width=0.35\textwidth]{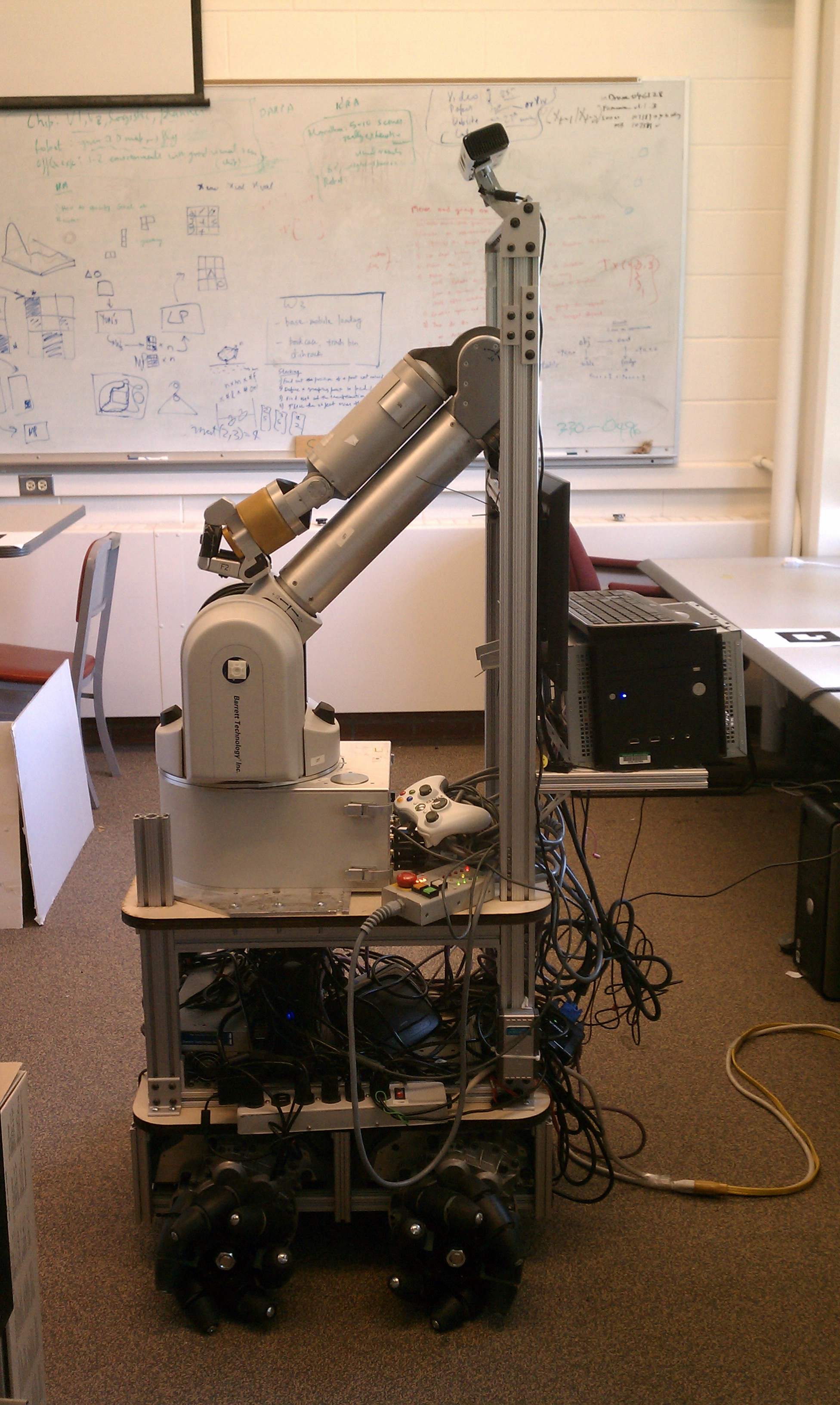} 
 \caption{Cornell's POLAR Robot. It consists of a omni-directional mobile base with a
 robotic arm and Kinect sensor mounted on the top.}
\label{fig:polarRobot}
 \end{figure}
   
   In this section, we first describe our dataset and present results of
    the scene labeling task on this dataset.  We present a detailed analysis 
    of various factors such as the effect of the 3D features, the effect of adding 
    context, the effect of the presence of unlabeled segments, etc. 
    
    We then 
    present the results on robotic experiments on attribute learning
    and contextually-guided    object detection. The object detection experiments were performed 
    using two robots: Cornell's Blue and POLAR robot, shown in Fig.~\ref{fig:blueRobot} and 
    Fig.~\ref{fig:polarRobot} respectively.  
    These robots consist of a mobile base (Erratic
    and Segway Omni 50 respectively) and were mounted with Kinect sensors as shown in respective figures.    (For more details on the hardware specification of the POLAR robot, please see \citealp{jiang2012placingobjects}.)
    We used ROS libraries\footnote{\url{http://www.ros.org}}  to program these robots.

   \subsection{Data}
 
  We consider labeling object segments in full 3D scene
(as compared to 2.5D data from a single view).
  For this purpose, we collected data of 24 office and 28 home scenes from a total
  of about 550 views. Each
scene was reconstructed from about 8-9 RGB-D views from a Microsoft
Kinect sensor and contains about one million colored points.
We first over-segment the 3D scene (as described earlier)
to obtain the atomic units of our representation.
For training, we manually labeled the segments, and we selected 
the labels which were present in
a minimum of 5 scenes in the dataset.  
Specifically, the office labels are: \{\emph{wall, floor, tableTop, tableDrawer,
tableLeg, chairBackRest,  chairBase, chairBack, monitor,
printerFront, printerSide keyboard,  cpuTop, cpuFront, cpuSide, book, paper}\},
and the home labels are: \{\emph{wall, floor, tableTop, tableDrawer, tableLeg, chairBackRest, 
 chairBase,  sofaBase, sofaArm, sofaBackRest, bed, bedSide, quilt, pillow, shelfRack, laptop, book}\}. This gave us a total of 1108 labeled segments in the office scenes and 1387 segments in the home scenes. 
 Often one object may be divided 
into multiple segments because of over-segmentation.
We have made this data available at: \url{http://pr.cs.cornell.edu/sceneunderstanding/data/data.php}.



\subsection{Segmentation}
For segmenting the point cloud, we use a region growing algorithm
similar to the Euclidean clustering in the Point-Cloud Library 
(PCL).\footnote{\url{http://www.pointclouds.org/documentation/tutorials/cluster_extraction.php}}
It randomly picks a seed point and grows it into a segment.
 New points are added to an existing segment if their distance
  to the closest point in 
 the segment is less than a threshold and the local normals calculated at these points are 
 at an angle less than a threshold.
 We also made the distance threshold proportional to the distance from camera because points
 far from the camera have more noisy depth estimates.
In detail, we used a distance threshold of $0.1d$, where $d$ is the distance of the candidate point from camera, 
and an angle threshold of $30^\circ$. We observed that these thresholds slightly over-segmented almost all 
scenes into object-parts of desired granularity.
Such a simple approach would not perfectly segment all objects, but we find that
our learning method is quite robust to such noisy segmentation.

\begin{table*}[t!]
\caption{{\bf Learning experiment statistics.} The table shows average micro precision/recall, and average macro precision and recall for home and office scenes. }
 \vskip -.2in
 \label{tbl:overall_result}
{\footnotesize
\newcolumntype{P}[2]{>{\footnotesize#1\hspace{0pt}\arraybackslash}p{#2}}
\setlength{\tabcolsep}{2pt}
\centering
\resizebox{\hsize}{!}
 {
\begin{tabular}
{p{0.20\linewidth}p{0.16\linewidth}|P{\centering}{14mm}P{\centering}{12mm}P{\centering}{12mm}|P{\centering}{14mm}P{\centering}{12mm}P{\centering}{12mm} }\\
\whline{1.1pt} 
& & \multicolumn{3}{c|}{Office Scenes} & \multicolumn{3}{c}{Home Scenes}  \\
\cline{3-8}
& & \multicolumn{1}{c}{micro} & \multicolumn{2}{c|}{macro} & \multicolumn{1}{c}{micro} &  \multicolumn{2}{c}{macro}   \\
\whline{0.4pt} 
     features &  algorithm & $P/R$ & Precision  & \multicolumn{1}{c|}{Recall} &  $P/R$ & Precision &  \multicolumn{1}{c}{Recall}  \\ 
\whline{0.8pt} 
None &  \emph{max\_class} &  26.33 & 26.33 & 5.88 & 29.38 & 29.38 & 5.88\\
\whline{0.6pt} 
Image Only &           \emph{svm\_node\_only}                        & 46.67  & 35.73 & 31.67 &   38.00 & 15.03 & 14.50\\
Shape Only &              \emph{svm\_node\_only}                        & 75.36  & 64.56 & 60.88 &    56.25 & 35.90 & 36.52 \\
Image+Shape &         \emph{svm\_node\_only}                         & 77.97  & 69.44 & 66.23 &   56.50 & 37.18 & 34.73 \\
\whline{0.6pt} 
Image+Shape \& context & \emph{single\_frames}                 & 84.32 & 77.84 & 68.12 & 69.13 & 47.84 & 43.62 \\

\whline{0.6pt} 
Image+Shape \& context &  \emph{svm\_mrf\_assoc}   					& 75.94  & 63.89 & 61.79 &    62.50 & 44.65 & 38.34\\
Image+Shape \& context &  \emph{svm\_mrf\_nonassoc}   					& 81.45  &76.79  &70.07   & 72.38  & 57.82  & 53.62 \\
Image+Shape \& context &  \emph{svm\_mrf\_parsimon}	     			& 84.06  & 80.52  & 72.64   & 73.38  & 56.81  &54.80 \\

\whline{1.1pt} 
\end{tabular}
}
}
\end{table*}

  \begin{figure*} [t!]  
 \centering
 \includegraphics[height=3in, width=3.35in]{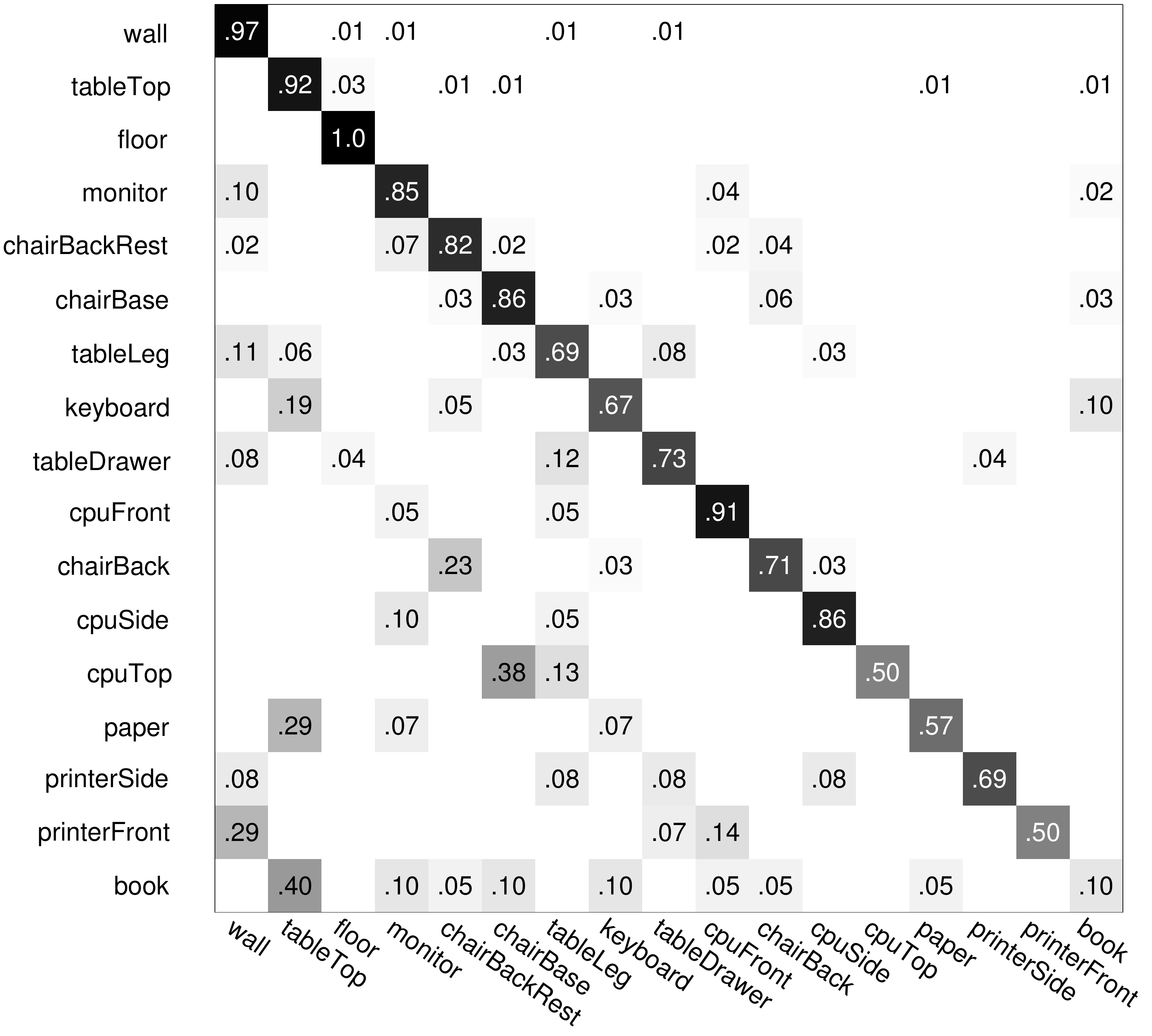} 
 \centering
 \includegraphics[height=3in,width=3.55in]{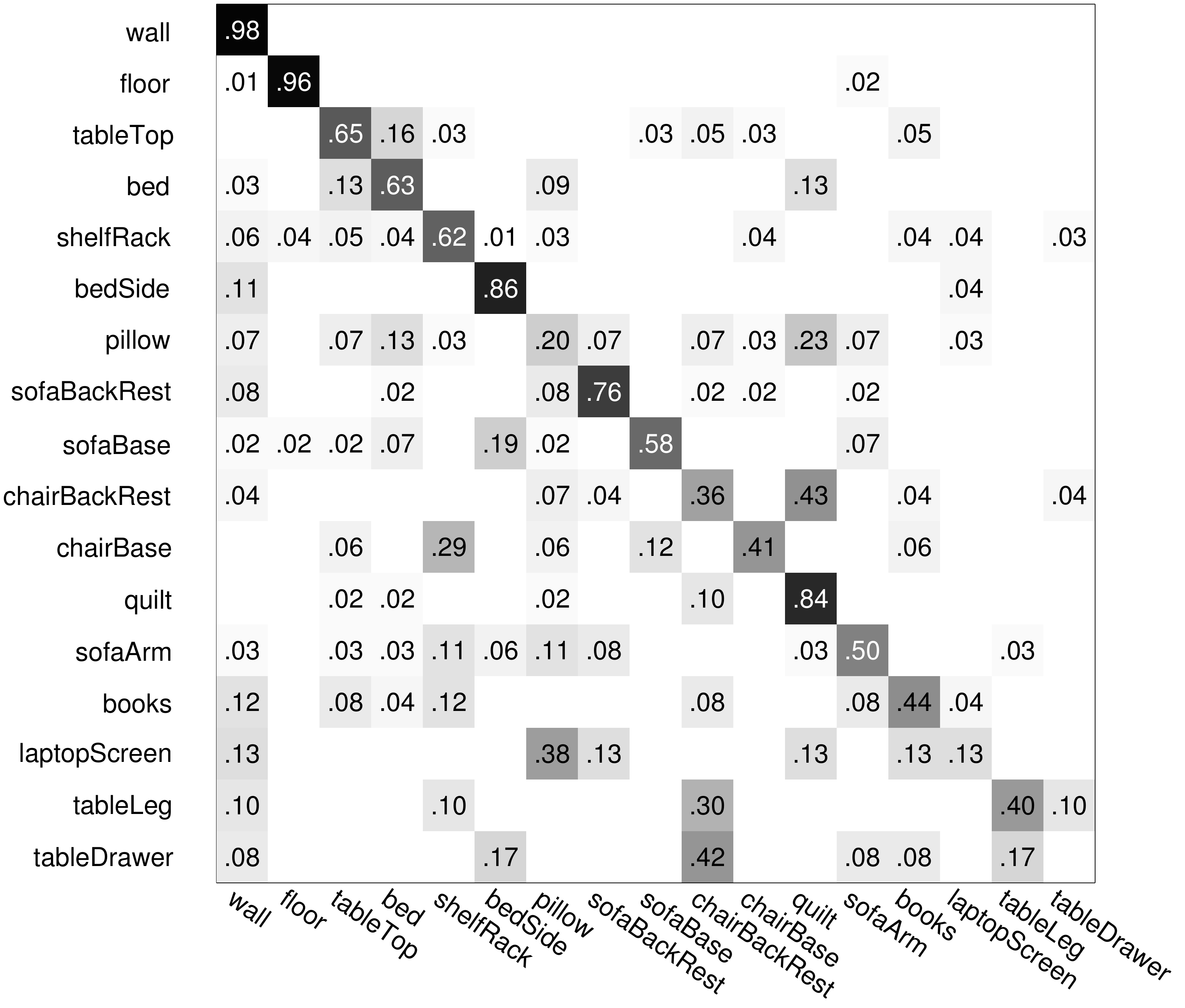} 
 \caption{Confusion Matrix on office dataset (left)  and home dataset (right) with \emph{svm\_mrf\_parsimon} trained on Shape and Image features.
}

\label{fig:confusionMatrix}
 \end{figure*}

   \subsection{Object Labeling Results}
\label{sec:labelingresults}

In this subsection, we report the results of offline labeling experiments.
Table \ref{tbl:overall_result} shows the results, performed using 4-fold
cross-validation and averaging performance across the folds for the models trained 
separately on home and office datasets.  
We use both the macro and micro averaging to aggregate precision and recall
over various classes.
Let $pred_i(k)$ denote that the $i^{th}$ segment was predicted to be of class $k$, and $gt_i(k)$ denote that the ground truth label of the $i^{th}$ segment was  class $k$. We now formally define the following metrics used in literature.
{\small
\begin{eqnarray*}
precision(k) &=& \frac{|\{i:(pred_i(k) \wedge gt_i(k))\}|}{|\{i:pred_i(k)\}|} \\
recall(k) &=& \frac{|\{i:(pred_i(k) \wedge gt_i(k))\}|}{|\{i:gt_i(k)\}|}\\
macro\_precision &=& \frac{\sum_{k=1}^{K}precision(k)}{K}\\
macro\_recall &=& \frac{\sum_{k=1}^{K}recall(k)}{K}\\
micro\_precision &=& \frac{\sum_{k=1}^{K} |\{i:(pred_i(k) \wedge gt_i(k))\}|}{\sum_{k=1}^{K} |\{i:pred_i(k)\}|}\\
micro\_recall &=& \frac{\sum_{k=1}^{K} |\{i:(pred_i(k) \wedge gt_i(k))\}|}{\sum_{k=1}^{K} |\{i:gt_i(k)\}|}\\
\end{eqnarray*}
}
where, for any set $S$, $|S|$ denotes its size.

In these experiments, prediction is done using an MIP solver with the constraint that a segment can have exactly one label ($\forall i: \sum_{j=1}^{K} \ysc{i}{j} = 1$).
 So,  micro precision and recall are same as the percentage of correctly
classified segments. 
The optimal C value is determined separately for each of the algorithms by
cross-validation. 
 

Fig.~\ref{fig:examplePCD} shows the original point cloud, ground-truth and predicted labels for one 
office (top) and one home scene (bottom). Fig.~\ref{fig:confusionMatrix} show the confusion matrices for office and home scenes on the left and right respectively. 
On majority of the classes our model predicts the correct label as can be seen from the strong 
diagonal in the confusion matrices. 
Some of the mistakes are reasonable, such as a pillow getting confused with the quilt in homes. They often have similar location and texture.
In offices, books placed on table-tops sometimes get confused with the table-top.

One of our goals is to study the effect of various factors, and therefore
we compared 
our algorithm with various settings.  
We discuss them in the following.
 
\medskip
\noindent \textbf{Do Image and Point Cloud Features Capture Complimentary Information?}
The RGB-D data contains both image and depth information, and enables us to
compute a wide variety of features.
In this experiment, we compare 
the two kinds of features: Image (RGB) and Shape (Point Cloud)
features.
To show the effect of the features independent of the effect of
context, we only use the node potentials from our model, referred to 
as \emph{svm\_node\_only}  in Table~\ref{tbl:overall_result}. The
\emph{svm\_node\_only} model is equivalent to the multi-class SVM formulation
\citep{joachims2009cutting}.  Table \ref{tbl:overall_result} shows that Shape
features are more effective compared to the Image, and 
the combination works better on both precision and recall.
This indicates that the two types of features offer complementary information and
their combination is better for our classification task.

\medskip
\noindent
\textbf{How Important is Context?}
Using our \emph{svm\_mrf\_parsimon} model as described in Section~\ref{sec:model},
we show significant improvements in the performance over using \emph{svm\_node\_only}
model on both datasets. In office scenes, the micro precision increased by
6.09\% over the best \emph{svm\_node\_only} model that does not use any context. 
In home scenes the increase is much higher, 16.88\%.

The type of contextual relations we capture depend on the type of 
edge potentials we model.
To study this, we compared our method with models using only associative or only
non-associative edge potentials referred to as \emph{svm\_mrf\_assoc} and
\emph{svm\_mrf\_nonassoc}.
We observed that modeling all edge features using associative potentials 
is poor compared to our full model. In fact, using only associative potentials
showed a drop in performance compared to \emph{svm\_node\_only} model on the office
dataset.  This indicates it is important to capture the relations between
regions having different labels.
Our \emph{svm\_mrf\_nonassoc} model does so, by modeling all edge features using
non-associative potentials, which can favor or disfavor labels
of different classes for nearby segments. It gives higher precision and recall
compared to \emph{svm\_node\_only} and \emph{svm\_mrf\_assoc}. This shows that modeling using
non-associative potentials is a better choice for our labeling problem.

However, not all the  edge features are  non-associative in nature, modeling
them using only non-associative potentials could be an overkill (each
non-associative feature adds $K^2$ more parameters to be learnt). Therefore
using our \emph{svm\_mrf\_parsimon} model to model these relations achieves higher
performance in both datasets.  



 \begin{figure} [h!]

 \includegraphics[scale=0.7]{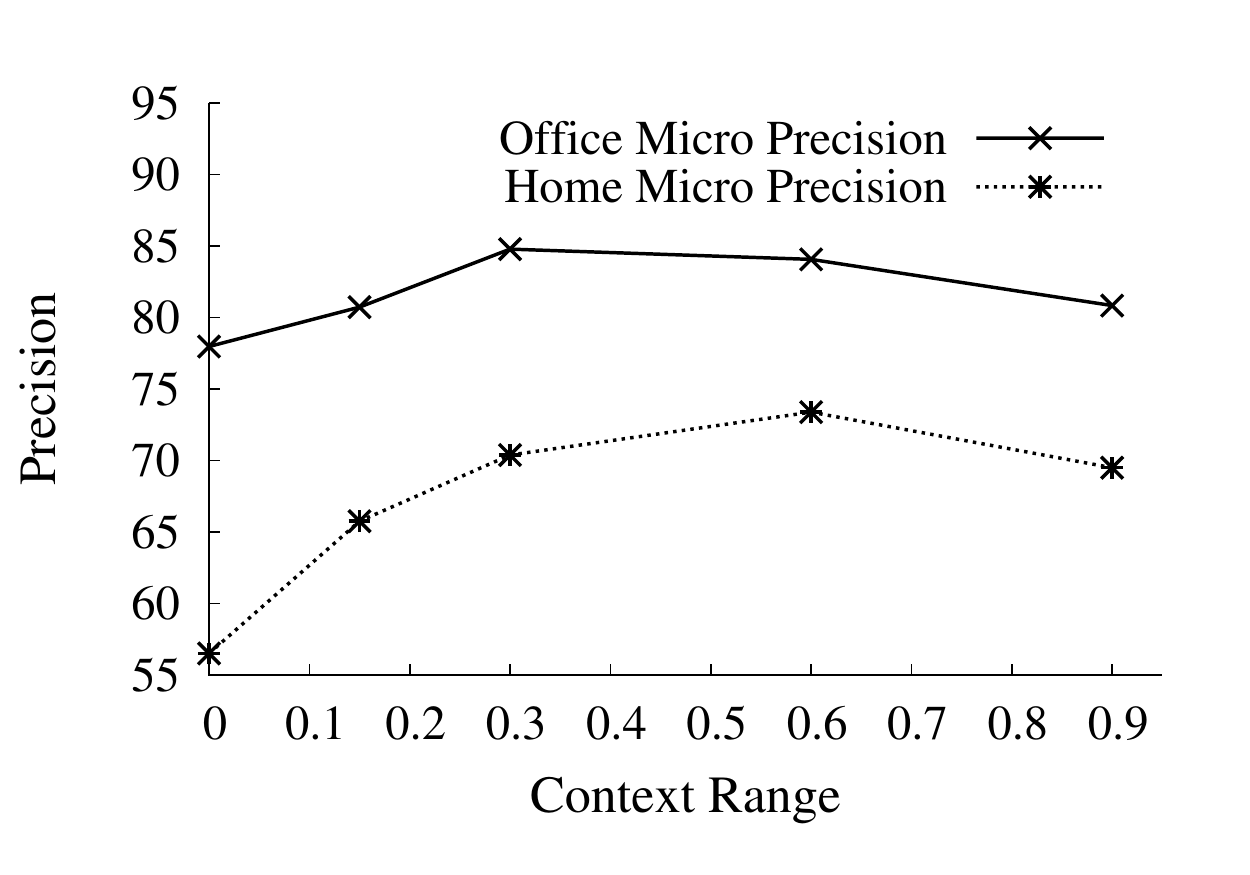} 
 \vskip -.1in
 \caption{Effect of context range on precision (=recall here).}
 \label{fig:radiusplot}
 \end{figure}

\medskip
\noindent 
\textbf{How Large should the Context Range be?} 
Context relationships of different objects can be meaningful for
different spatial distances.  This range may vary depending
on the environment as well. For example, in an office, keyboard and
monitor go together, but they may have little
relation with a sofa that is slightly farther away.  In a house, 
sofa and table may go together even if they are farther away.

In order to study this, we compared our \emph{svm\_mrf\_parsimon} with varying
$context\_range$ for determining the neighborhood (see Figure~\ref{fig:radiusplot}
for average micro precision vs range plot). Note that the $context\_range$ is determined
from the boundary of one segment to the boundary of the other, and hence it is
somewhat independent of the size of the object.  We note that increasing the
$context\_range$ increases the performance to some level, and then it drops slightly.
We attribute this to the fact that with increasing the $context\_range$, irrelevant objects
may get an edge in the graph, and with limited training data, spurious
relationships may be learned.  We observe that the optimal $context\_range$
for office scenes is around 0.3 meters and 0.6 meters for home scenes.



\medskip
\noindent
\textbf{How does a Full 3D Model Compare to a 2.5D Model?}
In Table~\ref{tbl:overall_result}, we compare the performance of our full model
with a model that was trained and tested on single-view point clouds of the same
scene. During the comparison, the training folds were consistent with other
experiments, however the segmentation of this point cloud was different (because
the input point cloud is from a single view). This makes the micro precision values 
meaningless because the distribution of labels is not same for the two cases. In particular, 
many large object in a scene (e.g., wall, ground) get split up into 
multiple segments in single-view point clouds. 
We observed that the macro precision and recall are higher when multiple views 
are combined to form the scene. 
We attribute the improvement in macro precision and recall to the fact that 
larger scenes have more context, and models are more complete because of multiple
views.



\medskip
\noindent 
\textbf{What is the effect of the inference method?}
The results for \emph{svm\_mrf} algorithms in Table \ref{tbl:overall_result} were generated using the MIP solver. 
 The QPBO algorithm however, gives a higher precision and lower recall on both datasets. For example, on office data, the graphcut inference for our \emph{svm\_mrf\_parsimon} gave a micro precision of 90.25 and micro recall of 61.74.  Here, the micro precision and recall are not same as some of the segments might not get any label.

%
%

\medskip
\noindent 
\textbf{What is the effect of having different granularity when defining the object classes?}
In our experiments, we have considered class labels at object-part level, e.g., classes \emph{chairBase}, \emph{chairBack} and \emph{chairBackRest} which are parts of a chair.
 We think that such finer knowledge of objects is important for many robotic applications. For example, if a robot is asked to arrange chairs in a room, 
knowing the chair parts can help determine the chair's orientation.  
Also, labeling parts of objects gives our learning 
algorithm an opportunity to exploit relationships between different parts of an object. 
In order to analyze the performance gain obtained by considering object-part level labeling, 
we compare our method with one trained on object level
classes. With 10 object level classes in office scenes : \emph{\{wall, table, chair, floor, cpu, book, paper, keyboard, printer and monitor\}},
we observe a drop in performance,
obtaining a micro precision/recall of 83.62\%, macro precision of 76.89\% and 
recall of 69.81\%.

   \begin{figure}  
 \includegraphics[width=1\linewidth]{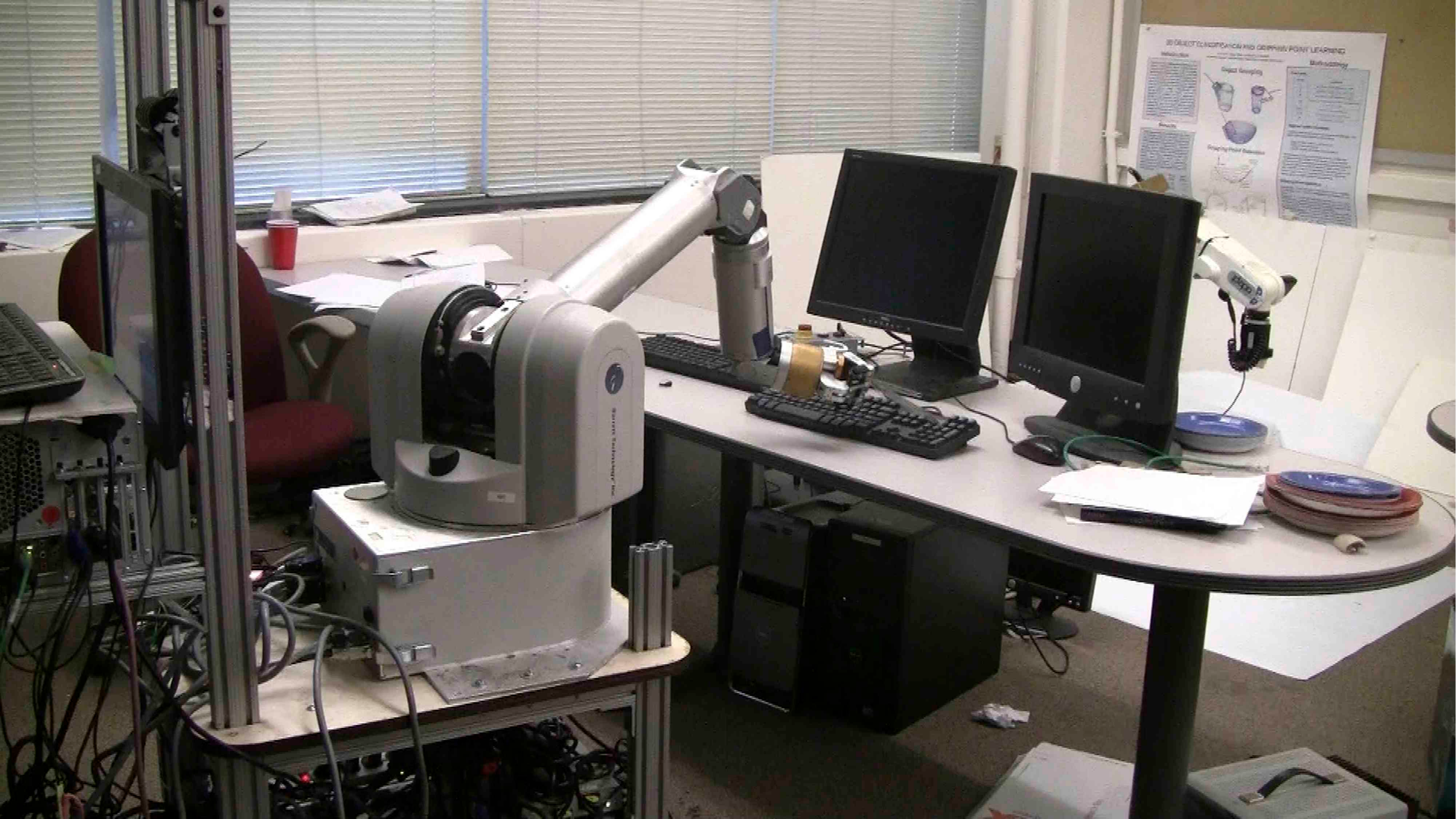} 
 \caption{Cornell's POLAR (PersOnaL Assistant Robot) using our classifier for detecting a keyboard in a cluttered room.}
 \vskip -.1in
 \label{fig:robot}
 \end{figure}

   \subsection{Attribute Labeling Results}
   
   In some robotic tasks, such as
robotic grasping, it is not necessary to know the exact object
category, but just knowing a few attributes of an object may
be useful. For example, if a robot has to clean a floor, it would
help if it knows which objects it can move and which it cannot.
If it has to place an object, it should place them on horizontal
surfaces, preferably where humans do not sit. With this motivation we have
designed 8 attributes, each for the home and office scenes, giving a total of 10
unique attributes. They are: \{\emph{wall, floor, flat-horizontal-surfaces,
furniture, fabric, heavy, seating-areas, small-objects, table-top-objects,
electronics}\}. Note that each segment in the point cloud can have multiple
attributes and therefore we can learn these attributes using our model which
naturally allows multiple labels per segment.  We compute the precision 
and recall over the attributes by counting how many attributes were correctly
inferred. In home scenes we obtained a precision of 83.12\% and
70.03\% recall, and in the office scenes we obtain 87.92\%
precision and 71.93\% recall.
   
   \subsection{Robotic Experiments}
   \label{sec:robotExps}
The ability to label scenes is very useful for robotics applications,
such as of 
finding/retrieving an object on request. 
As described in Section~\ref{sec:computingPredictions}, in a detection scenario there
can be some segments not belonging to the object classes we consider. 
Table~\ref{tbl:detection_results} shows the results of running our inference algorithms for detection scenario
on the offline office dataset when considering all segments, including those belonging to classes other than the 17 mentioned earlier.
The solution of the relaxed MIP (described in Section~\ref{sec:computingPredictions}) gives us 
 high precision  (89.87\% for micro,
and 82.21\% for macro), but low recall (55.36\% for micro, and 35.25\% for macro). 
The $\hat{\y}_{cut}$ solution, computed using the graph-cut method, also achieves comparable accuracy (see line 3 of Table \ref{tbl:detection_results}) and is very fast to compute (takes less than 0.05 second per scene). 
Therefore, if the robot finds an object it is 
likely correct, but the robot may 
not find all the objects easily. This is where our contextual search algorithm (described in 
Section~\ref{sec:contextualSearch}) becomes useful.




\begin{table}[h]
\caption{Precision and Recall for detection experiments in office scenes (offline, single-view).}
\begin{center}
 \label{tbl:detection_results}
\begin{tabular}{l | c | c | c | c}
\whline{1.1pt} 
 Algorithm & Micro-  & Micro-  & Macro-  & Macro-\\
  &  precision &  recall &  precision &  recall\\
\whline{0.8pt} 
 \emph{max\_class} & 22.64 & 22.64 & 22.64 & 5.88 \\
 \emph{svm\_mrf\_parsimon}  & 89.87 & 55.36 & 82.21 & 35.25 \\
 \emph{svm\_mrf\_parsimon} & \multirow{2}{*}{87.41} & \multirow{2}{*}{56.82} & \multirow{2}{*}{82.95} & \multirow{2}{*}{38.14} \\
 \emph{$\quad$w/ QPBO, $\hat{y}_{cut}$}   & & & &\\
\whline{1.1pt} 
\end{tabular}
  
  \end{center}
\end{table}

  
\begin{table}
  \caption{Class-wise precision recall for Robotic Experiments using contextual search. 
 }
\begin{center}
 \label{tbl:robotic_exp_result}
\begin{tabular}{l | c | c | c}
\whline{1.1pt} 
class & \# instances & precision & recall \\
\whline{0.8pt} 
Wall & 10 & 100 & 100 \\
Table Top  & 10 & 100  &100  \\
Table Leg & 10 & 71 & 50  \\
Table Drawer & 7 & 100 & 71 \\
Chair Backrest & 10 & 100 & 100  \\
Chair Base & 10 & 100  & 100  \\
Chair Back & 8 & 100 & 88  \\
Monitor & 9 & 100 & 100  \\
Keyboard & 9 & 100 & 78 \\
CPU & 8 & 50 & 11  \\ 
Printer & 1 & 100 & 100 \\
Paper & 10 & 100 & 22 \\
\whline{0.6pt} 
Overall Micro &\multirow{2}{*}{102} & 96 & 75 \\
Overall Macro & & 93 & 77 \\
\whline{1.1pt} 
\end{tabular}
  \end{center}
  \vskip -.2in
\end{table}

\begin{figure*}[t!]

\includegraphics[width=0.19\linewidth,height=1in]{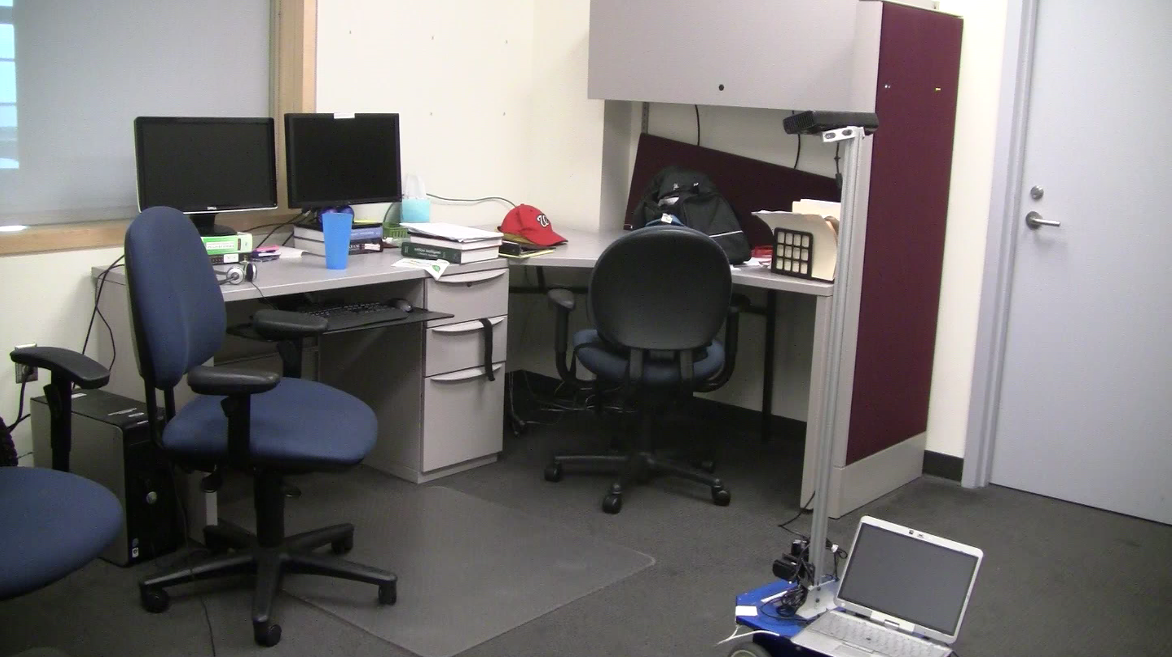}
\includegraphics[width=0.19\linewidth,height=1in]{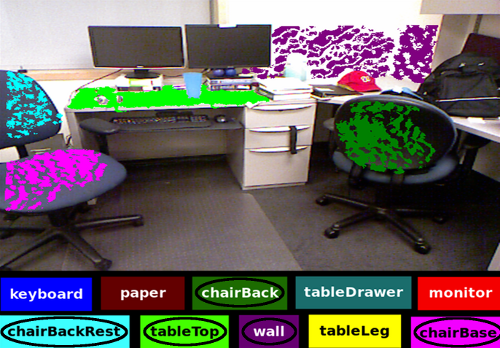} 
\includegraphics[width=0.19\linewidth,height=1in]{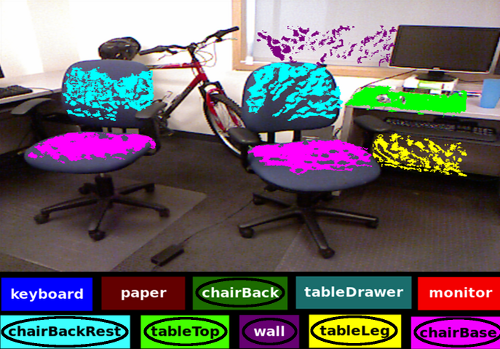} 
\includegraphics[width=0.19\linewidth,height=1in]{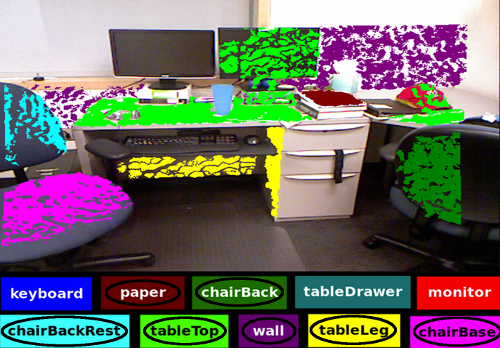} 
\includegraphics[width=0.19\linewidth,height=1in]{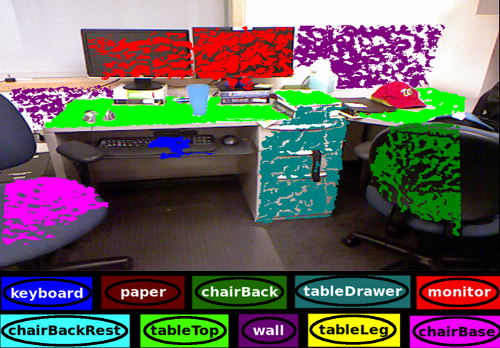} 
\vskip -.1in
 \caption{(Left):  The robot in an office scene. (Columns 2-5): Sequence of colored images corresponding to the labeled point clouds generated by the robot during the object detection experiment.}
\label{fig:robotExpt}
 \end{figure*}

    \begin{figure*}[t!]
 \centering
 \includegraphics[width=0.95\linewidth,height=0.25in]{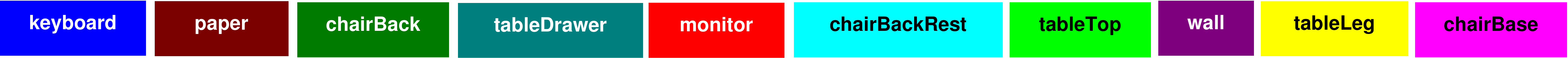}\\
 \vskip 0.04in
\includegraphics[width=0.32\linewidth]{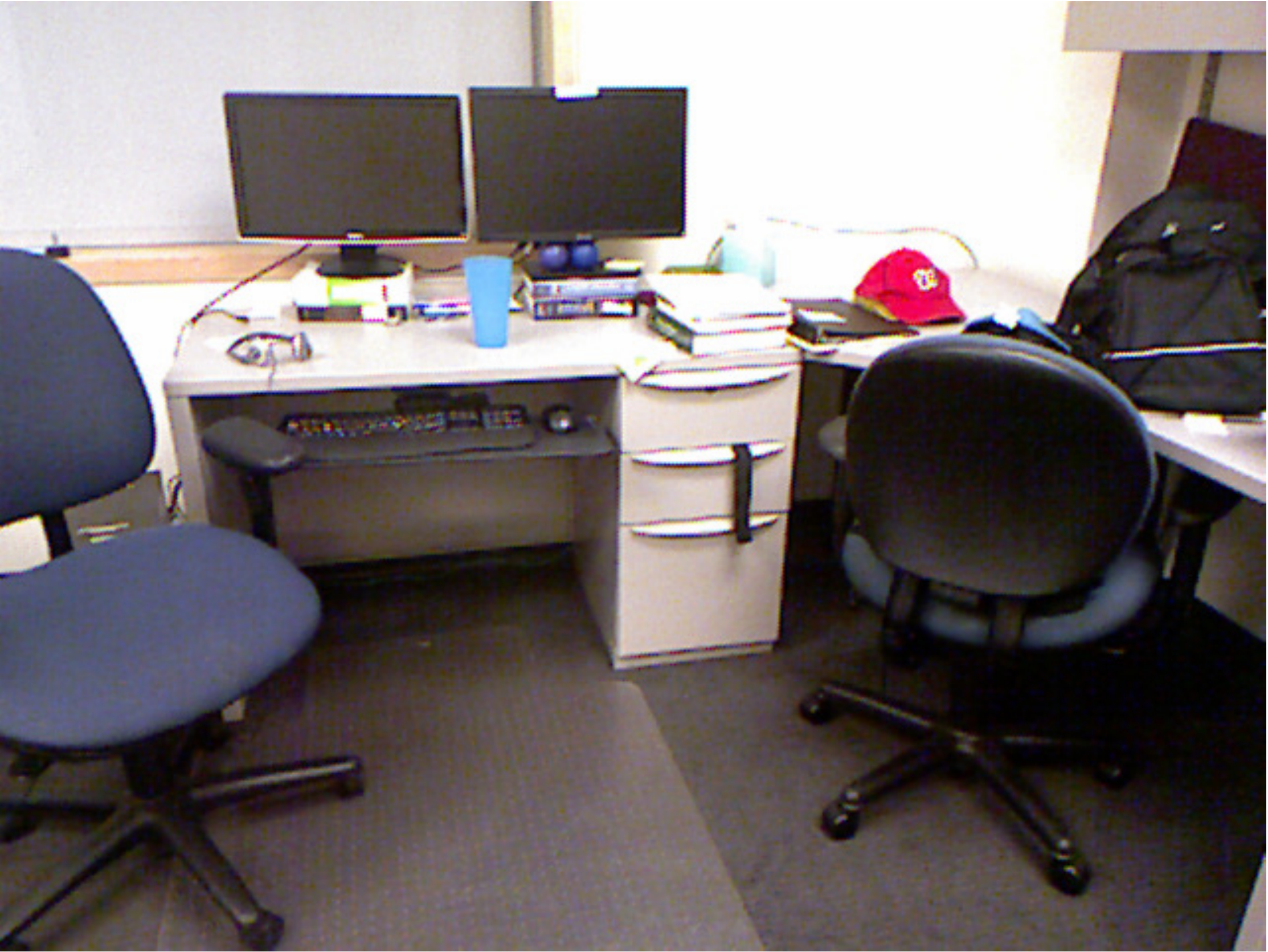} 
\includegraphics[width=0.32\linewidth]{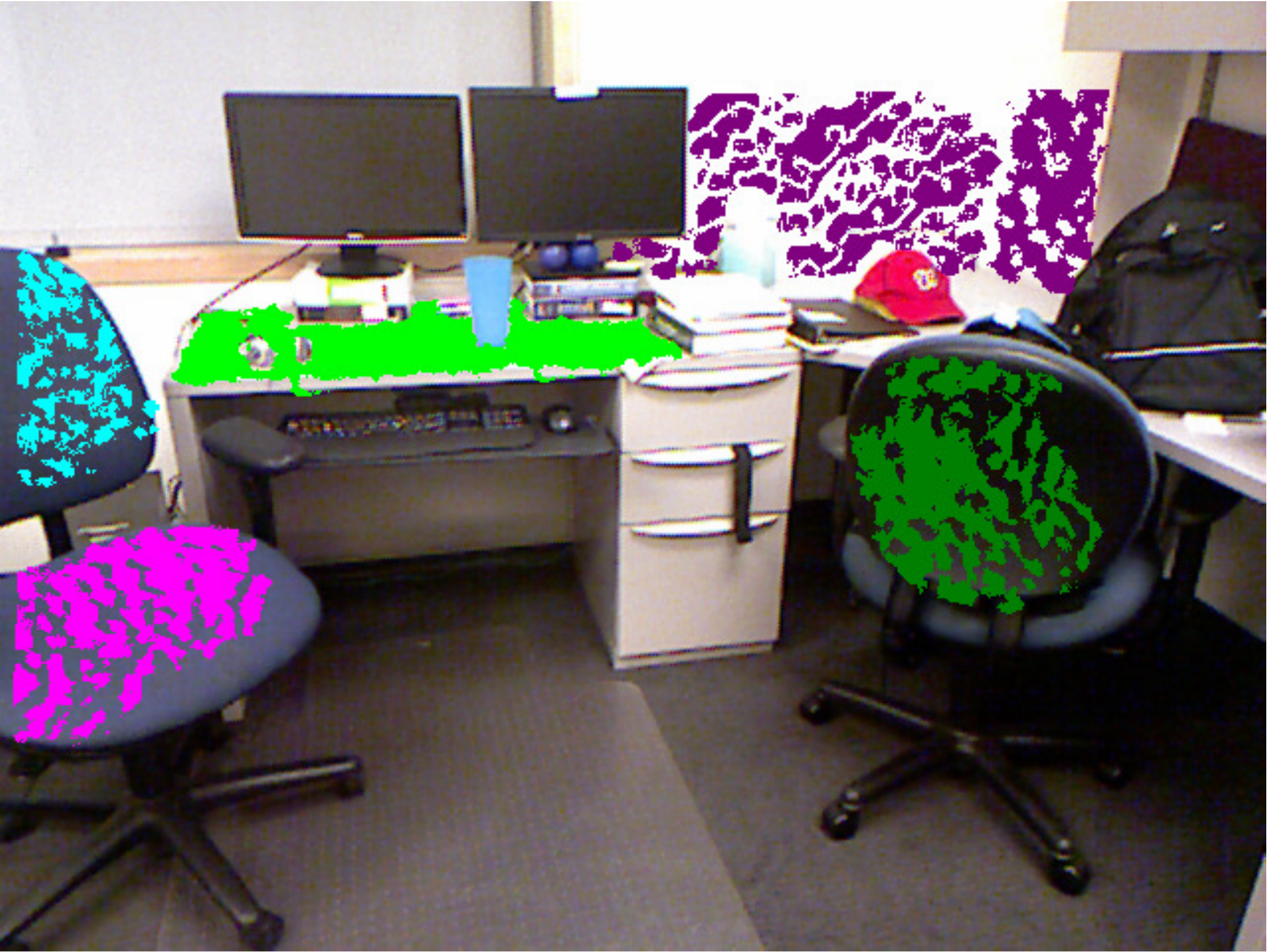} 
\includegraphics[width=0.32\linewidth]{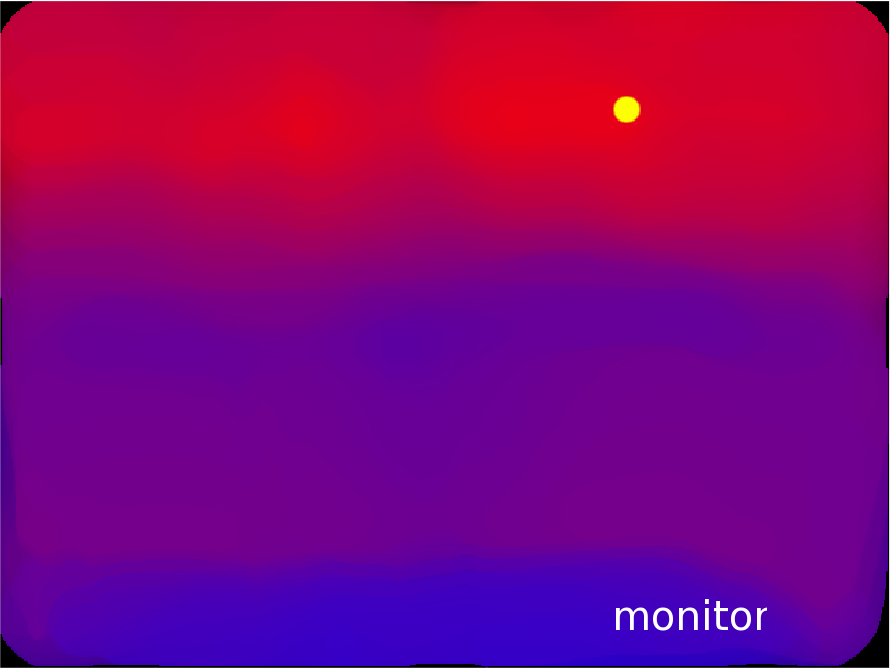} \\
\vskip 0.04in
\includegraphics[width=0.32\linewidth]{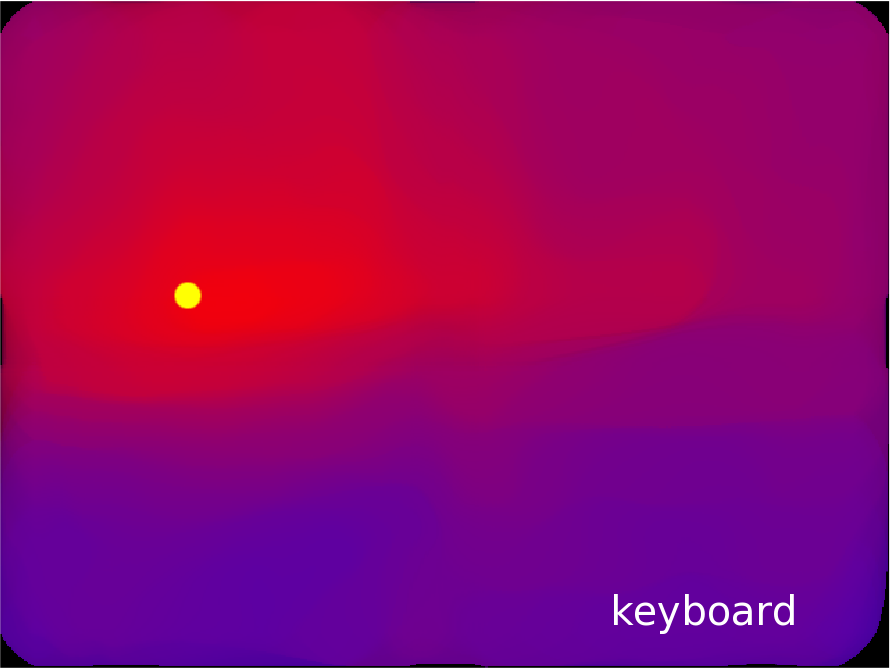} 
\includegraphics[width=0.32\linewidth]{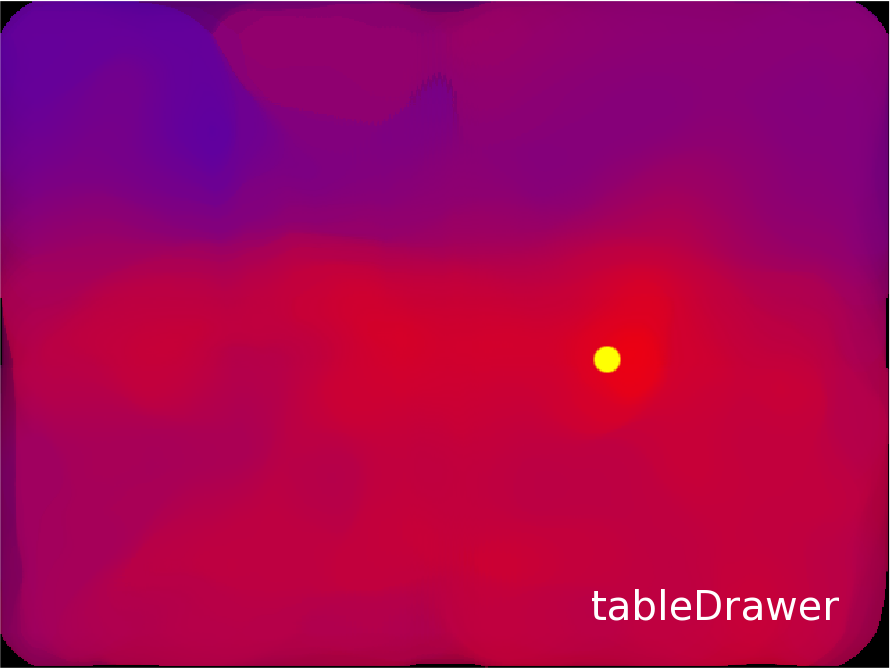} 
\includegraphics[width=0.32\linewidth]{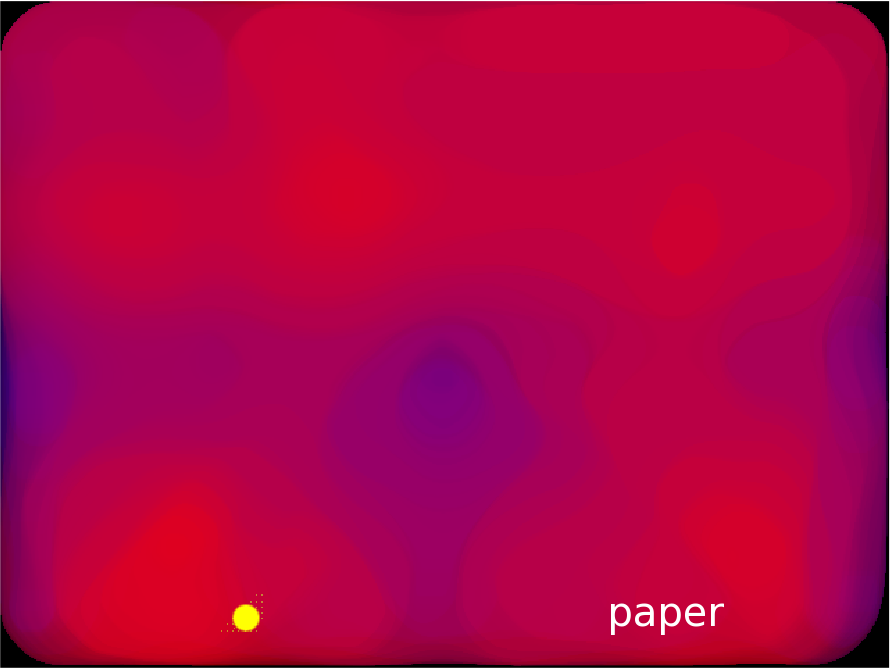} 
 \caption{(Top left) Original image. (Top mid) Inferred labels color-coded using the legend shown above. (Top right, bottom left, bottom mid and bottom right): Contextually likely positions for finding monitor, keyboard, tableDrawer and paper respectively. In these heatmaps, red indicates that the target object is more likely to be found there. The circular yellow dot indicates the most likely location.}
\label{fig:heatmaps}
\vskip -.15in
 \end{figure*}

\begin{figure*}
\vskip -.05in
\includegraphics[width=0.5\linewidth,height=2in]{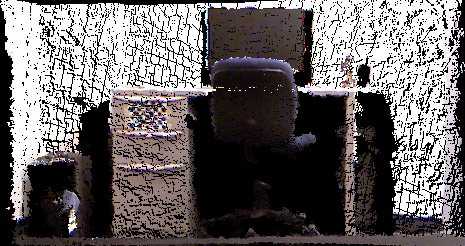}
\includegraphics[width=0.5\linewidth,height=2in]{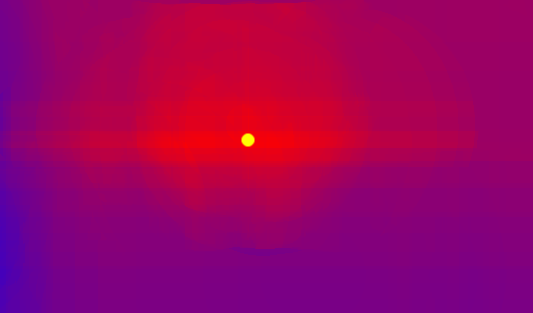} \\
\includegraphics[width=0.5\linewidth,height=2in]{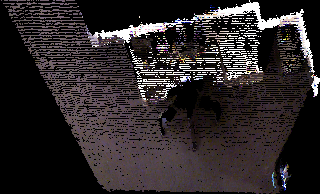}
\includegraphics[width=0.5\linewidth,height=2in]{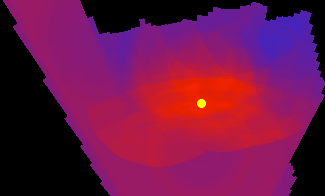} 
 \caption{(Left column): Front and top views(orthographic projections) of an office scene in which a keyboard is occluded. (Right column): Corresponding heatmaps where red indicates that the keyboard is more likely to be found there. The yellow dot indicates the most likely location.}
\label{fig:heatmapsOcc}
\vskip -.15in
 \end{figure*}

In order to evaluate our contextual search algorithm, we test our approach on our Blue robot for the task of finding 12 object classes 
located in 10 office scenes.  The robot starts from a pre-determined location in an office and searches for 
a given list of objects in the room. The goal of the robot is to find at least 
one instance for each of the object classes it is asked to search for. Since, the RGB-D sensor
has a narrow field-of-view (57 degrees horizontally),  
the robot first scans the room by turning a fixed angle each time. It labels the 
point cloud it obtains in each view and saves the labeled point clouds. 

Next, for all the object classes it did not find, it computes the contextually likely
locations for the objects using the algorithm described in Section~\ref{sec:contextualSearch}.
Using the the inferred locations, the robot moves in that direction 
in order to get better view of the objects. Fig.~\ref{fig:robotExpt} shows the 
experiment run in one office scene. The first column shows the Blue robot and the scene in which 
it is asked to find the objects and the rest of the columns show the sequence of 
colored images corresponding to the labeled point clouds. The first two point clouds were 
obtained when the robot was scanning the room from a distance, and the last two are 
obtained after inferring the contextually likely locations of the objects not found and moving 
closer to these locations.  
Table \ref{tbl:robotic_exp_result} shows the precision and recall of finding the 12 object classes
in our robotic experiments.

To evaluate our contextual search algorithm, we present both qualitative and quantitative results.
Fig.~\ref{fig:heatmaps} qualitatively shows the predictions for finding monitor, keyboard, tableDrawer and paper in a frame in which they were not found. 
These heatmaps are generated in 3D, but for visualization purposes they are aligned 
to the original RGB image in Fig.~\ref{fig:heatmaps}.
As can be seen form Fig.~\ref{fig:heatmaps}, a monitor is predicted to be most likely found above the table and a tableDrawer is likely to be found under the table. However, it does not do a great job for paper, and this may be because we had very few examples for this class in our training set.
 
Fig.~\ref{fig:heatmapsOcc} shows that our algorithm can also be used on a robot to find objects even when they are occluded. Clearly, it predicts that a keyboard is likely to be found in front of the monitor and at about the same height as the tableTop.
To quantitatively evaluate the predictions, we collected 10 frames where a keyboard was not detected, but other objects such as table and monitor were detected.  We then applied our contextual search algorithm to 
find the optimal 3D locations for finding a keyboard. For each of these scenes, we computed the minimum distance of the actual keyboard-points to the inferred optimal location. 
As a baseline, if the predictor always predicted midpoints of the scene as the probable location of the keyboard, the max, mean and median values over 10 scenes were  $113.5$cm, $32.6$cm and $27.2$cm respectively.
Using our method, we get $36.3$cm, $17.5$cm and $15.9$cm respectively.  We found that this helps the robot in finding the objects with only a few moves.


We have the code available as a ROS and PCL package.
Code, datasets as well as videos showing our robots finding objects using our algorithm are available at:
   \url{http://pr.cs.cornell.edu/sceneunderstanding/}.

\section{Conclusion}

In conclusion, we have proposed and evaluated the first model and learning 
algorithm for semantic labeling that exploits rich relational information
in full-scene 3D point clouds. Our method captures 
various features and contextual relations, including the local visual appearance and shape cues, object co-occurence relationships and geometric relationships.
We showed how visual and shape features can be modeled parsimoniously when the number of classes is large.
We also presented an algorithm to infer contextually likely locations for the desired objects
given the current labelings in the scene.
We tested our method on a large offline dataset, as well as on the task of mobile robots
finding objects in cluttered scenes.


\section{Acknowledgements}
The final version of this paper will be published in IJRR, 2012 by SAGE Publications Ltd, All rights reserved.
We thank Gaurab Basu, Yun Jiang, Jason Yosinski and Dave Kotfis for help with the robotic experiments. This research was funded in part by Microsoft Faculty Fellowship and Alfred P. Sloan Research Fellowship to one of us (Saxena), and by NSF Award IIS-0713483.

{  
\bibliographystyle{apalike}
\bibliography{references}
}

\end{document}